%% file: main.tex
\pdfoutput=1  

\documentclass[letterpaper, 10 pt, conference]{ieeeconf}  

\IEEEoverridecommandlockouts                              

\input{tex/preamble_packages}
\input{tex/shortcuts.tex}

\crefname{Appendix}{Appendix}{Appendices}
\Crefname{Appendix}{Appendix}{Appendices}

\title{\LARGE \bf
3D Scene Graph Prediction: Generating Hierarchical Models from Partially Observed Environments
}

\author{Siyi Hu$^{1}$, Jared Strader$^{1,2}$, Hyungtae Lim$^{1}$, and Luca Carlone$^{1}$
\thanks{*This work was partially funded by the ARL DCIST program, the ONR RAPID program, and National Research Foundation of Korea (No. RS-2024-00461409).}
\thanks{$^{1}$Laboratory for Information \& Decision Systems (LIDS), Massachusetts Institute of Technology, Cambridge, MA 02139, USA. {\tt\small \{siyi, jstrader, shapelim, lcarlone\}@mit.edu}}%
\thanks{$^{2}$Department of Electrical and Computer Engineering, Oakland University, Rochester, MI 48309, USA. {\tt\small jstrader@oakland.edu}}%
}

\begin{document}
\maketitle
\thispagestyle{empty}
\pagestyle{empty}

\input{tex/abstract}

\glsresetall    
\input{tex/introduction}
\input{tex/relatedWork}
\input{tex/preliminary}
\input{tex/approach}
\input{tex/experiment}
\input{tex/conclusion}

\bibliographystyle{plain}
\bibliography{references/refs,references/refs_HSG} 

\clearpage

\appendix
\crefalias{section}{Appendix}
\crefalias{subsection}{Appendix}
\crefalias{subsubsection}{Appendix}
\input{tex/supplementary/implementation}
\input{tex/supplementary/post_processing}
\input{tex/supplementary/hydra_processing}
\input{tex/supplementary/additionalResults}
\end{document}

%% file: tex/preamble_packages.tex
\usepackage{comment}
\usepackage{siunitx}
\usepackage{relsize}
\usepackage{ifthen}

\usepackage[vlined,ruled,linesnumbered]{algorithm2e}
\usepackage{graphics} 
\usepackage{rotating}
\usepackage{color}
\usepackage{enumerate}
\usepackage[T1]{fontenc}
\usepackage{psfrag}
\usepackage{epsfig} 
\usepackage{subcaption}
\usepackage{booktabs}
\usepackage{graphicx,url}
\usepackage{multirow}
\usepackage{array}
\usepackage{latexsym}
\usepackage{amsfonts}
\usepackage{amsmath}
\usepackage{amssymb}
\usepackage{xstring}
\usepackage[noend]{algorithmic}
\usepackage{multirow}
\usepackage{xcolor}
\usepackage{hyperref}
\usepackage[nameinlink,noabbrev]{cleveref}
\usepackage{prettyref}
\usepackage{flexisym}
\usepackage{bigdelim}
\usepackage{breqn} 
\usepackage{listings}
\let\labelindent\relax
\usepackage{enumitem}
\usepackage{xspace}
\usepackage{bm}
\usepackage{placeins}
\graphicspath{{./figures/}}
\usepackage{tikz}
\usetikzlibrary{matrix,calc}

\usepackage{mdwlist}

\let\itemize\undefined
\makecompactlist{itemize}{stditemize}

\usepackage[acronym]{glossaries}
\glsdisablehyper

%% file: tex/shortcuts.tex
\newcommand{\myParagraph}[1]{{\bf #1.}\xspace}
\newcommand{\etal}{\emph{et~al.}\xspace}
\newcommand{\eg}{\emph{e.g.,}\xspace}
\newcommand{\ie}{\emph{i.e.,}\xspace}

\newacronym{mp3d}{MP3D}{Matterport3D}
\newacronym{bro}{BRO}{building--room--object}
\newacronym{3dsg}{3DSG}{3D Scene Graph}
\newacronym{ssc}{SSC}{Semantic Scene Completion}
\newacronym{gnn}{GNN}{Graph Neural Network}
\newacronym{wl}{WL}{Weisfeiler-Lehman}
\newacronym{vae}{VAE}{Variational Autoencoder}
\newacronym{gan}{GAN}{Generative Adversarial Network}
\newacronym{llm}{LLM}{Large Language Model}
\newacronym{vlm}{VLM}{Vision Language Model}
\newacronym{mlp}{MLP}{Multi-Layer Perceptron}
\newacronym{ddpm}{DDPM}{Denoising Diffusion Probabilistic Model}
\newacronym{d3pm}{D3PM}{Discrete Denoising Diffusion Probabilistic Model}
\newacronym{sde}{SDE}{Stochastic Differential Equation}
\newacronym{esdf}{ESDF}{Euclidean Signed Distance Function}
\newacronym{sgd}{SGD}{Stochastic Gradient Descent}
\newacronym{miqp}{MIQP}{mixed-integer quadratic program}
\newacronym{tsp}{TSP}{Travelling Salesman Problem}
\newacronym{fid}{FID}{Fr\'{e}chet Inception Distance}
\newacronym{kid}{KID}{Kernel Inception Distance}
\newacronym{kl}{KL}{Kullback-Leibler}
\newacronym{mmd}{MMD}{maximum mean discrepancy}
\newacronym{emd}{EMD}{Earth Mover's Distance}

\newcommand{\M}[1]{{\bm #1}}

\newcommand{\E}{{\mathbb{E}}}   
\newcommand{\calN}{{\cal N}}

\newcommand{\tran}{^{\mathsf{T}}}
\newcommand{\MI}{\M{I}}         
\newcommand{\Real}[1]{ { {\mathbb R}^{#1} } }
\newcommand{\Int}[1]{ { {\mathbb Z}^{#1} } }

\newcommand{\ContinuousVar}{\M{x}}

\newcommand{\discreteVar}{z}
\newcommand{\discreteDomain}{{\cal Z}}

\newcommand{\oneHotVec}[1]{\M{v}( #1 )}

\newcommand{\DiscreteTransMat}{\M{Q}}

\newcommand{\semanticVar}{z}
\newcommand{\edgeVar}{e}
\newcommand{\nodeDim}{D_{\ContinuousVar}}   

\newcommand{\graph}{G}
\newcommand{\numNodes}{N}
\newcommand{\nodes}{{\cal V}}
\newcommand{\edges}{{\cal E}}

\newcommand{\CondVar}{\M{y}}
\newcommand{\networkVar}{\theta}
\newcommand{\partialk}{\text{partial}}
\newcommand{\known}{\text{known}}
\newcommand{\unknown}{\text{unknown}}

\newcommand{\numLabels}{C}

\newcommand{\PosVar}{\M{t}}
\newcommand{\ShapeVar}{\M{s}}

%% file: tex/abstract.tex
\begin{abstract}

Generating realistic 3D indoor scenes is an area of growing interest in computer vision and robotics.
Existing methods, often motivated by applications such as interior design, generally focus on object layout generation within a single room.
The generation of high-level scene structure, such as room-level layout and traversability, remains underexplored despite its importance for robotics applications.
In this paper, we consider the case where a robot has explored part of an environment and needs to predict the unexplored parts to support downstream tasks such as exploration or object search.
We propose a top-down framework for synthesizing hierarchical 3D scene graphs, including a room layer---describing the floor plan and traversability---and an object layer modeling object layouts within each room.
For the room layer, we propose a novel mixed-domain graph diffusion model jointly predicting room categories, floor boundaries, and traversability between rooms.
Via corruption and masking, this model supports partial constraints such as incomplete floor plans, avoiding the need for partially observed training data.
For the object layer, we integrate an existing mixed discrete-continuous diffusion model for joint prediction of object categories, locations, sizes, and orientations within each room given the floor plan.
We compare our method with state-of-the-art occupancy-based and LLM-based floor plan generation methods on a standard benchmark.
Compared with an occupancy-based learning baseline, our method generalizes substantially better to out-of-distribution partial floor plans.
We also demonstrate our integrated prediction pipeline on real-world scenes from robot-collected data, enabling prediction beyond explored areas.

\end{abstract}

%% file: tex/introduction.tex

\section{Introduction}
Scene generation and prediction play a crucial role in robotics and computer vision, enabling machines to understand, anticipate, and interact with their environments.
Advancements in deep learning and graph-based methods have significantly improved scene understanding, enabling autonomous systems to reason about spatial relationships and predict plausible scene configurations to support tasks such as exploration, navigation, and object search.

\Glspl{3dsg}, particularly \emph{hierarchical} scene graphs~\cite{Armeni19iccv-3DsceneGraphs,Hughes24ijrr-hydraFoundations} that represent 3D scenes at different levels of abstraction (\eg objects, rooms), have been widely adopted in robotics as a lightweight representation of 3D environments. The hierarchy provides organization for scene understanding and reasoning~\cite{Hughes24ijrr-hydraFoundations}.
However, occlusions and partial observability are common in robotic exploration, making it challenging to predict what lies in unexplored regions and plan efficient exploration strategies.
Without the ability to reason about unobserved portions of the environment, robots cannot make informed decisions about where to explore next or anticipate the spatial structure beyond explored areas.
Moreover, scene completion is often hampered by the lack of training data due to heterogeneous sensing modalities and environmental conditions; 
collecting in-distribution data to cover all possible scenarios is often infeasible.

\input{figures/apartment/apartment_integration}

To address these challenges, we propose a hierarchical \gls{3dsg} prediction framework for indoor scenes where, given an observed scene graph at the current time, we predict the scene graph corresponding to the unobserved portion of the scene.
We consider a 3-layer scene graph, consisting of building, room, and object layers, and develop a top-down generation approach.
Our contributions are as follows:
\begin{itemize}
    \item We introduce a mixed-domain diffusion model for the room layer that jointly predicts latent geometric features, semantic labels, and inter-room connectivity. To handle partial observations at test time without retraining, we propose a masking strategy that enables flexible conditioning on missing inputs.
    \item We couple the room-layer predictor with MiDiffusion~\cite{Hu26wacv-midiffusion}, an object-layer predictor conditioned on room geometry and type, yielding an end-to-end pipeline for building-scale scene completion.
    \item We validate the proposed framework through extensive experiments on the 3D-FRONT and Matterport3D datasets, demonstrating improved performance over state-of-the-art methods on room-graph completion.
\end{itemize}
Implementation details and additional results are provided in the appendix.

%% file: figures/apartment/apartment_integration.tex

\begin{figure}[!t]
    \centering
    \begin{subfigure}[t]{0.5\linewidth}
        \vspace{0pt}
        \includegraphics[width=\linewidth,trim={10 20 10 0},clip]{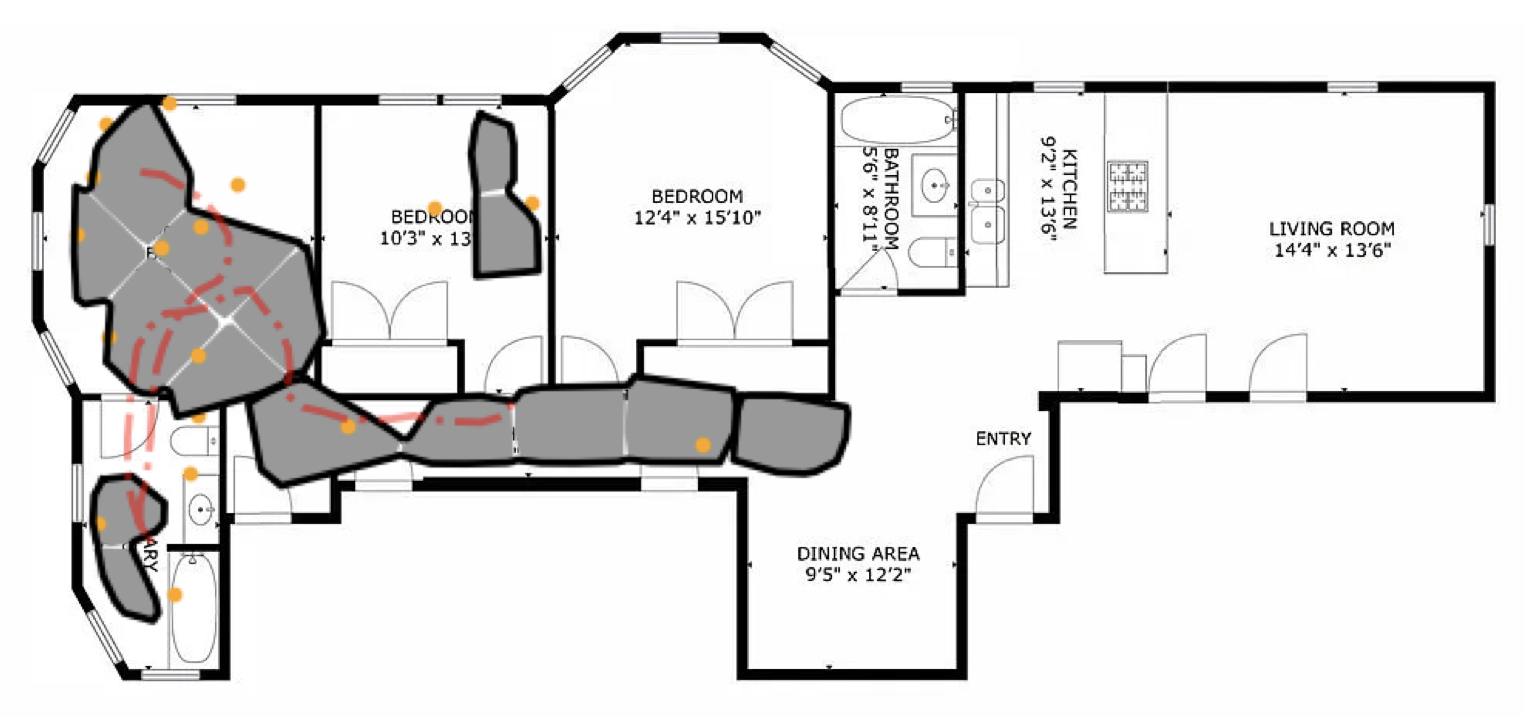}
        \caption{The observed portion of the scene, overlayed on a complete (unknown) floor plan.}
        \label{fig:apartment_partial}


        \hspace{-5pt}
        \includegraphics[width=0.94\linewidth,trim={50 180 0 180},clip]{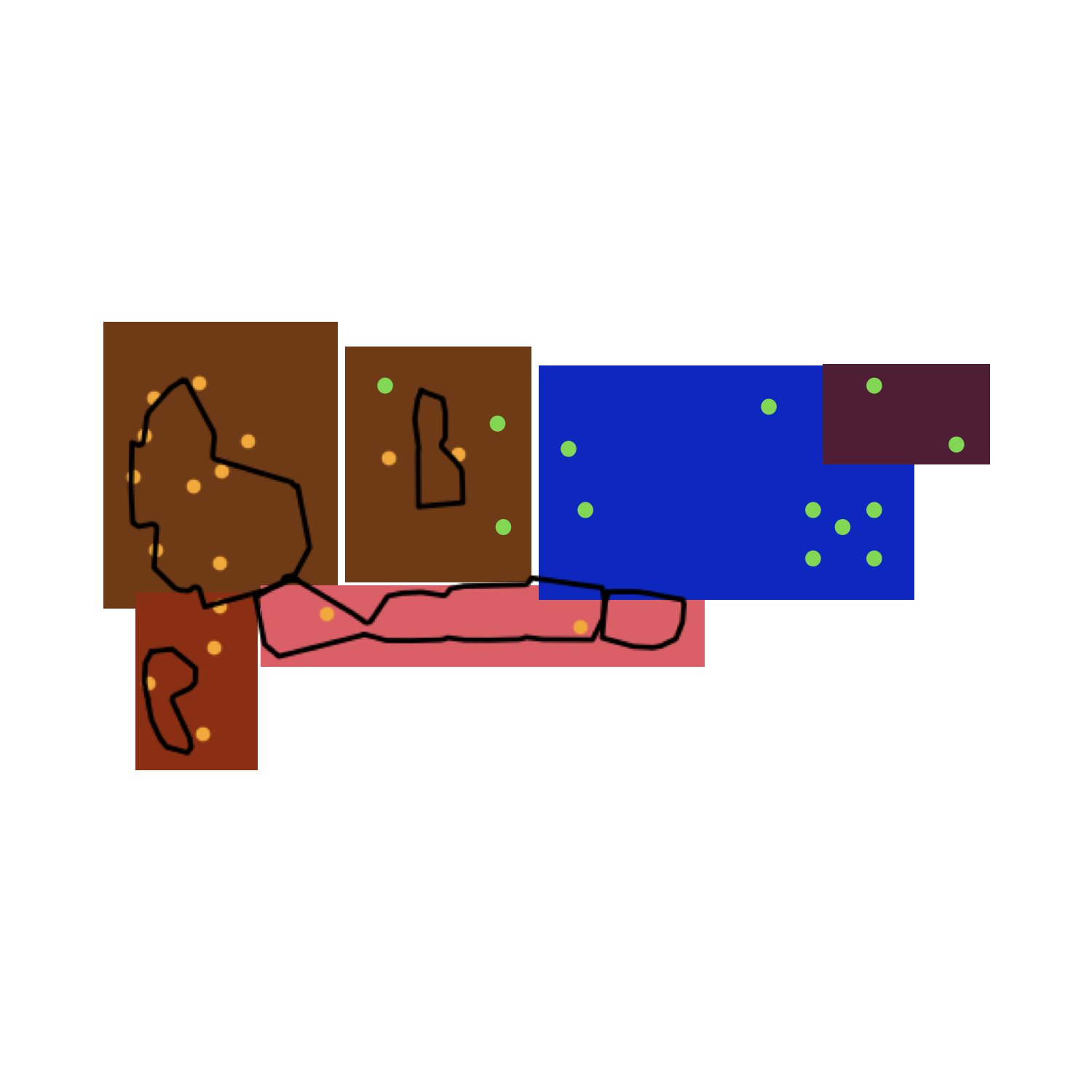}
        \vspace{-9pt}
        \caption{Completed floor plan from \;\; partial observations.}
        \label{fig:apartment_sc}
    \end{subfigure}\hfill
    \begin{subfigure}[t]{0.5\linewidth}
        \vspace{16pt}
        \centering
        \includegraphics[width=\linewidth,trim={40 0 00 0},clip]{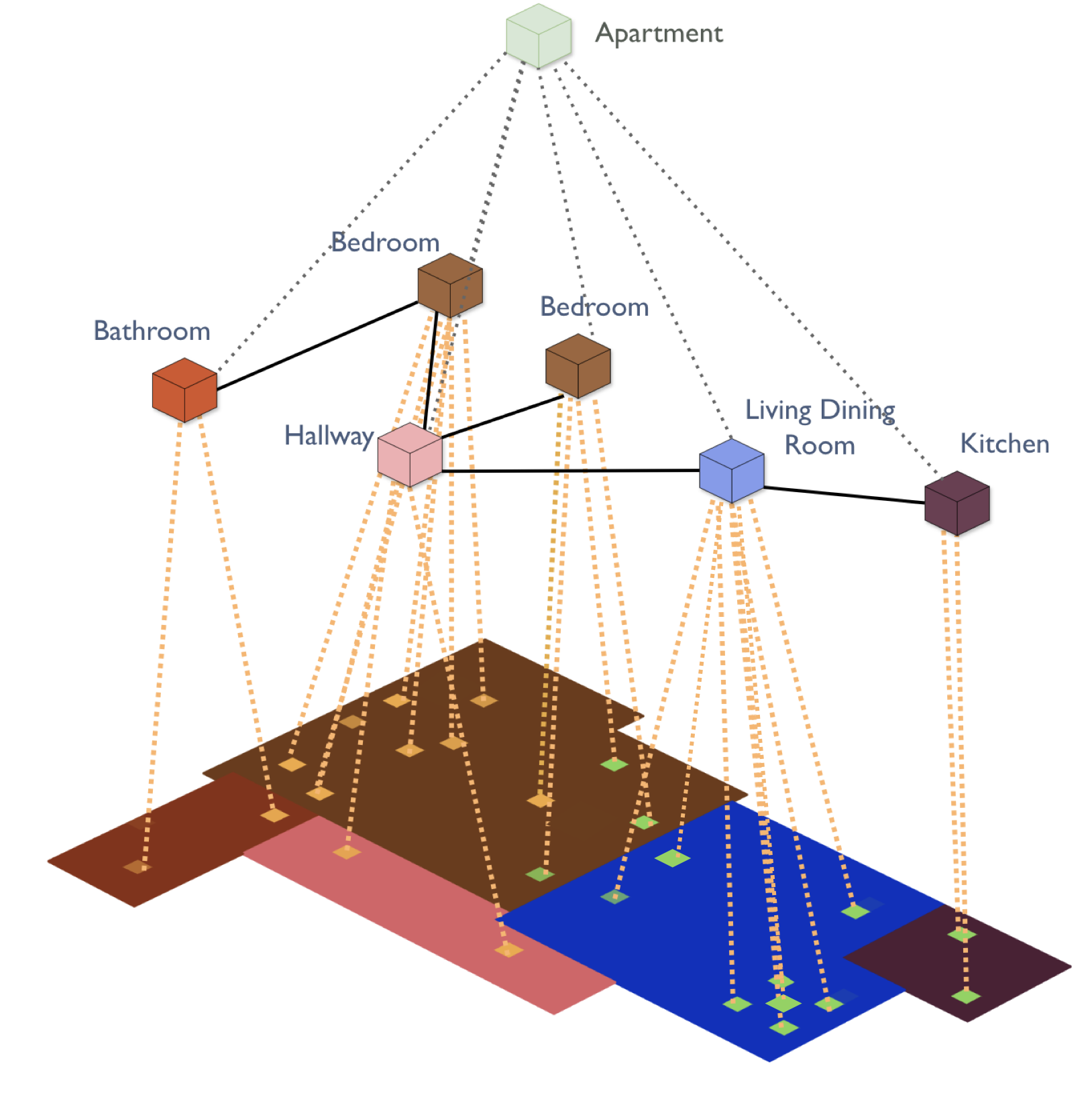}
        \caption{Layered view of the completed apartment scene.}
        \label{fig:apartment_sc_3d}
    \end{subfigure}

    \vspace{-2pt}
    \caption{Real-world hierarchical 3D scene graph prediction. 
    Gray patches denote the observed portion of the scene; colored polygons represent predicted rooms; orange dots indicate detected and green dots indicate synthesized objects.}
    \label{fig:apartment_integration}
    \vspace{-13pt}
\end{figure}

%% file: tex/relatedWork.tex
\section{Related Work}\label{sec:related_works}

\myParagraph{3D Scene Synthesis}
\Glspl{3dsg} provide a compact, hierarchical representation of indoor environments for robotics~\cite{Armeni19iccv-3DsceneGraphs,Hughes24ijrr-hydraFoundations}. 
Hydra~\cite{Hughes24ijrr-hydraFoundations} enables real-time construction of multi-layer scene graphs from sensor data, including free-space representations useful for navigation. Recent systems~\cite{Gu24icra-conceptgraphs,Maggio24ral-clio} incorporate open-vocabulary reasoning.

While \glspl{3dsg} offer a structured map, synthesizing such scenes remains challenging. Most prior work focuses on from-scratch prediction and is limited mainly to flat object layouts for single rooms.
Early methods use procedural modeling to encode object relationships~\cite{Yu11tog-furnitureArrangement, Devaranjan20eccv-MetaSim2}. Recent approaches learn inter-object relationships using \glspl{vae}~\cite{Yang21iccv-SceneSynthesis}, autoregressive models~\cite{Li19tog-GRAINS, Zhou19iccv-SceneGraphNet, Paschalidou21neurips-ATISS}, and diffusion models~\cite{Tang24cvpr-DiffuScene,Lin24iclr-instructScene,Hu26wacv-midiffusion}. \gls{llm}-based planners~\cite{Feng23neurips-LayoutGPT, Yang24cvpr-Holodeck, Fu24eccv-AnyHome} and text-driven video synthesis~\cite{Fridman23neurips-SceneScape} have also been explored.
However, only a few works address room-level floor plans~\cite{Nauata21cvpr-House-GAN++, Fu24eccv-AnyHome, Yang24cvpr-Holodeck}, typically with limited support for partial constraints.

\myParagraph{World Modeling and Scene Completion}
Beyond from-scratch synthesis, autonomous agents often need to predict unobserved regions from partial observations.
\Gls{ssc} for outdoor environments predicts missing semantic voxels or point clouds~\cite{Wang24neurips-SSC,Lee23arxiv-diffusionSSC, Lee24cvpr-SemCity}, often using dual-branch networks to handle partial inputs~\cite{Wang24neurips-SSC,Lee23arxiv-diffusionSSC}.
For indoor settings, recent works use symbolic representations for instance-level world modeling: Yin~\etal~\cite{Yin24neurips-SG-Nav} use scene graphs with \glspl{llm} for navigation, while Zhao~\etal~\cite{Zhao25iclr-ImagineNav} exploit the hierarchy of \glspl{3dsg} to prompt \glspl{llm} for structural priors. 
However, current \glspl{llm} and \glspl{vlm} can still struggle with complete scene prediction and precise geometry due to limited 3D spatial grounding.

\myParagraph{Diffusion Models for Layout Synthesis}
Diffusion models~\cite{Ho20neurips-DiffusionModel, Austin21neurips-D3PM} have been widely adopted for synthesis tasks, including document layouts~\cite{Inoue23cvpr-LayoutDM}, 3D scenes~\cite{Tang24cvpr-DiffuScene, Lin24iclr-instructScene}, graph-conditioned layouts~\cite{Zhai24eccv-EchoScene}, and 3D meshes~\cite{Gao24cvpr-GraphDreamer, Meng25cvpr-LT3SD}.
Following \cite{Sohl-Dickstein15icml-diffusion}, these models add noise through a forward process and denoise through a reverse process. \glspl{ddpm}~\cite{Ho20neurips-DiffusionModel} use Gaussian noise for continuous data, while discrete diffusion models~\cite{Austin21neurips-D3PM} operate on categorical data.
Most layout synthesis works use \glspl{ddpm} in continuous space for joint semantic and geometric inference~\cite{Tang24cvpr-DiffuScene, Lin24iclr-instructScene}. LayoutDM~\cite{Inoue23cvpr-LayoutDM} applies discrete diffusion to document layouts, and MiDiffusion~\cite{Hu26wacv-midiffusion} proposes a mixed discrete-continuous approach for 3D scenes.

\myParagraph{Generative Models for Graphs}
Graph generative models predict node and edge attributes, often using autoregressive architectures~\cite{You18icml-GraphRNNGenerating}, which struggle with equivariance.
Recent score-based and diffusion frameworks leverage equivariant denoising networks~\cite{Satorras21icml-En-gnn, Hoogeboom22icml-equivariantDiffusion} to jointly model node and edge distributions via systems of \glspl{sde}~\cite{Jo22icml-GDSS}.
Latent graph diffusion~\cite{Zhou24neurips-LatentGraphDiffusion} scales generation via compressed embeddings.
Discrete graph diffusion models~\cite{Vignac23iclr-DiGress, Chen23icml-EDGE} have shown strong performance in predicting discrete labels where continuous noise struggles to capture graph structure.
In 3D scene synthesis, graphs are primarily used for text instruction encoding~\cite{Lin24iclr-instructScene, Fu24eccv-AnyHome}, whereas our work focuses on direct generation of hierarchical graph structures for robotics.

%% file: tex/preliminary.tex
\section{Preliminaries: Diffusion Models}

Diffusion models~\cite{Sohl-Dickstein15icml-diffusion} define a forward noising Markov chain and a learned reverse denoising chain.
Given a data variable $\ContinuousVar_0$ (continuous or discrete), the forward process $q(\ContinuousVar_{1:T} \,|\, \ContinuousVar_0)=\prod_{t=1}^T q(\ContinuousVar_t \,|\, \ContinuousVar_{t-1})$ progressively corrupts $\ContinuousVar_0$ into $\ContinuousVar_T$, while the reverse process $p_\theta(\ContinuousVar_{0:T})=p(\ContinuousVar_T)\prod_{t=1}^T p_\theta(\ContinuousVar_{t-1}\mid \ContinuousVar_t)$ denoises by a network parameterized by $\theta$.
Training minimizes a variational bound $L_{vb}$ that decomposes into per-timestep terms:
\begin{equation}
\label{eq:diffusion_loss}
\begin{split}
    L_{vb} &=\mathbb{E}_{q(\ContinuousVar_0)}[
    \underbrace{D_{\mathrm{KL}}\left(q(\ContinuousVar_T | \ContinuousVar_0)|| p(\ContinuousVar_T)\right)}_{L_T} \\
    & +\sum_{t=2}^T \underbrace{\E_{q(\ContinuousVar_t|\ContinuousVar_0)}[D_{\mathrm{KL}}(q(\ContinuousVar_{t-1} | \ContinuousVar_t, \ContinuousVar_0)|| p_\theta(\ContinuousVar_{t-1} | \ContinuousVar_t))]}_{L_{t-1}} \\
    & + \underbrace{\E_{q(\ContinuousVar_1|\ContinuousVar_0)}[-\log p_\theta(\ContinuousVar_0 | \ContinuousVar_1)]}_{L_0}
    ].
\end{split}
\end{equation}
When $q$ injects known noise, $L_T$ is constant and can be dropped.
Diffusion models support conditional synthesis by conditioning the reverse chain on an input $\CondVar$ (omitted for brevity).
We first review continuous and discrete diffusion; in \Cref{sec:approach}, we combine them at the graph level.

\myParagraph{Continuous (Gaussian) diffusion}
DDPM~\cite{Ho20neurips-DiffusionModel} injects Gaussian noise into a continuous variable $\ContinuousVar_0$ with a fixed variance schedule $\beta_1, \dots, \beta_T \in (0, 1)$:
\begin{equation}
    \label{eq:forward_mc}
    q(\ContinuousVar_t | \ContinuousVar_{t-1}) := \calN(\ContinuousVar_t; \sqrt{1-\beta_t}\ContinuousVar_{t-1}, \beta_t \MI).
\end{equation}
This yields closed-form sampling $q(\ContinuousVar_t \,|\, \ContinuousVar_0)$ and tractable posteriors, enabling efficient training via Eq.~\eqref{eq:diffusion_loss}.

\myParagraph{Discrete (categorical) diffusion}
For discrete data, corruption is more naturally defined in the discrete domain.
We denote a scalar discrete variable with $K$ categories as $\discreteVar \in \Int{K}$ and use $\oneHotVec{\discreteVar_t} \in \{0, 1 \}^K$ for its one-hot encoding (the variational bound in Eq.~\eqref{eq:diffusion_loss} also holds for $\discreteVar$).
The forward process at time $t$ is defined by a transition probability matrix $\DiscreteTransMat_t$, with $[\DiscreteTransMat_t]_{mn} = q(\discreteVar_t =m | \discreteVar_{t-1}=n)$:
\begin{equation}
    \begin{split}
        q(\discreteVar_{t} | \discreteVar_{t-1}) & := \text{Cat}(\discreteVar_{t}; p=\DiscreteTransMat_t \oneHotVec{\discreteVar_{t-1}}) \\
        & = \oneHotVec{\discreteVar_t}\tran \DiscreteTransMat_t \oneHotVec{\discreteVar_{t-1}}.
    \end{split}
\end{equation}
The denoising network is trained to predict $p_\theta(\discreteVar_{t-1} \mid \discreteVar_t)$.
We adopt the mask-and-replace strategy of VQ-Diffusion~\cite{Gu22cvpr-VQ-Diffusion}, extending D3PM~\cite{Austin21neurips-D3PM} with an additional \texttt{[MASK]} token.

\myParagraph{Graph diffusion models}
Graph diffusion models apply the same framework to graph-structured data $\graph := (\nodes,\edges)$~\cite{Jo22icml-GDSS,Vignac23iclr-DiGress}, where node attributes $\nodes$ and edge attributes $\edges$ may be continuous, discrete, or mixed.
Prior work typically commits to either continuous score-based~\cite{Jo22icml-GDSS} or discrete label diffusion~\cite{Vignac23iclr-DiGress}; in \Cref{sec:approach}, we derive a mixed-domain formulation that injects domain-specific noise while learning a joint denoising network.

%% file: tex/approach.tex

\section{Approach}\label{sec:approach}

We assume a 3-layer hierarchical \gls{bro} scene graph $\graph = (\nodes, \edges)$, where nodes are partitioned into building, room, and object layers (\ie $\nodes=\nodes^B \uplus \nodes^R \uplus \nodes^O$).\footnote{$\uplus$ denotes disjoint union.}
We adopt the hierarchical graph definition from Hydra~\cite{Hughes24ijrr-hydraFoundations}: each node in the room and object layers has one parent in the layer above (\ie each object belongs to a room, each room to a building), and nodes share edges only with nodes in adjacent layers (\ie object nodes do not connect to building nodes). Thus, edges are partitioned as $\edges=\edges^{BR} \uplus \edges^{R} \uplus \edges^{RO} \uplus \edges^{O}$.

Due to the geometric nature of \glspl{3dsg}, the inter-layer edges $\edges^{BR}$ and $\edges^{RO}$ can be inferred directly from geometric features (\eg an object lying within a room boundary). 
We therefore adopt a top-down approach that synthesizes the room-level scene graph first and then the object-level graphs. For an indoor multi-room environment, this decomposes hierarchical scene graph completion into two sub-problems: room-level scene graph generation and object-level scene graph generation conditioned on the parent-room geometry.
We consider two settings: unconditional generation from scratch and completion from a partial scene graph.

\subsection{Room-level Scene Graph Generation}
We assume the room-level scene graph $\graph^R = (\nodes^R, \edges^R)$ contains room nodes, each characterized by a semantic label $\semanticVar^{R,i}\in \discreteDomain = \{1, 2, \dots, \numLabels \}$ and a geometric feature $\ContinuousVar^{R,i} \in \Real{\nodeDim}$, along with undirected edges $\edgeVar^{R,ij} \in \{0,1\}$ indicating traversability between rooms $i$ and $j$.
We assume at most $\numNodes$ rooms, so $\graph^R = (\{( \semanticVar^{R,i}, \ContinuousVar^{R,i})\}_{1 \leq i \leq \numNodes}, \{ \edgeVar^{R,ij} \}_{1 \leq i < j \leq \numNodes})$.
The geometric feature combines the floor plan center $\PosVar^{R,i}\in\Real{3}$ and a 2D shape feature $\ShapeVar^{R,i}\in\Real{\nodeDim-3}$, a latent-space encoding of the room shape, \ie $\ContinuousVar^{R,i}=(\PosVar^{R,i}, \ShapeVar^{R,i})$.
We designate the last semantic label $\semanticVar^{R}=\numLabels$ as an ``empty'' label to handle scenes with fewer than $\numNodes$ rooms, setting geometric attributes to zero for ``empty'' rooms.
For simplicity, we drop the superscript $R$ for the remainder of this section.
We next describe our mixed-domain graph diffusion formulation, the denoising network architecture, and our strategy for conditioning on partial observations.

\subsubsection{Graph Diffusion in Mixed Domains}
The room graph generation problem involves predicting variables in three domains: $\graph = \graph^{\semanticVar} \times \graph^{\ContinuousVar} \times \graph^{\edgeVar}$.
We derive a graph diffusion framework as a general mixed-domain diffusion process, circumventing the domain conversion step in existing works.

Following Eq.~\eqref{eq:diffusion_loss}, synthesizing a graph $\graph_0$ via diffusion amounts to minimizing
\begin{equation}
\label{eq:graph_variational_bound}
\resizebox{.91\linewidth}{!}{$\displaystyle
\begin{aligned}
    &L_{vb} = 
    \mathbb{E}_{q(\graph_0)}\left[ 
    \sum_{t=2}^T \underbrace{\E_{q(\graph_t|\graph_0)}[D_{\mathrm{KL}}(q(\graph_{t-1} | \graph_t, \graph_0)\,||\, p_\theta(\graph_{t-1} | \graph_t))]}_{L_{t-1}} \right. \\
    &\qquad \qquad + \left. \underbrace{\E_{q(\graph_1|\graph_0)}[-\log p_\theta(\graph_0 | \graph_1)]}_{L_0}\right]
\end{aligned}
$}
\end{equation}
by dropping the constant $L_T$ term from Eq.~\eqref{eq:diffusion_loss}.
If we assume $\graph$ is a product of mixed-domain variables, $\graph=\prod_{k}\graph^k$ (in our case, $k \in \{\semanticVar, \ContinuousVar, \edgeVar\}$), and we inject independent domain-specific noise, the forward process factorizes as:
\begin{equation}
    \label{eq:graph_forward_process}
    q(\graph_{t} | \graph_{t-1}) := \prod_{k} q^k(\graph^k_{t} | \graph^k_{t-1}).
\end{equation}
Sampling from the forward process and computing the posterior are then also domain-independent:
\begin{equation}
    \label{eq:graph_forward_sample}
    q (\graph_{t} | \graph_{0}) = \prod_{k} q^k(\graph^k_{t} | \graph^k_{0}),
\end{equation}
\begin{equation}
    \label{eq:graph_forward_posterior}
    q (\graph_{t-1} | \graph_{t}, \graph_{0}) = \prod_{k} q^k(\graph^k_{t-1} | \graph^k_{t}, \graph^k_{0}).
\end{equation}
For the backward process, we design a network $\networkVar$ to output domain-specific distributions of $\graph_{t-1}$ given $\graph_t$:
\begin{equation}
    \label{eq:graph_backward_process}
    p_\networkVar (\graph_{t-1} | \graph_{t}) := \prod_{k} p^k_\networkVar (\graph^k_{t-1} | \graph_{t}).
\end{equation}
Combining these, the losses in Eq.~\eqref{eq:graph_variational_bound} decompose into domain-specific terms:
\begin{subequations}
    \label{eq:graph_loss_decompose}
    \begin{align}
    &\resizebox{.88\linewidth}{!}{$\displaystyle L_{t-1} = \sum_{k} \underbrace{\E_{q^k(\graph^{k}_t|\graph^{k}_0)} \left[ D_{\mathrm{KL}} \left( q^k(\graph^{k}_{t-1} | \graph^{k}_{t}, \graph^{k}_{0}) \,||\, p^k_\networkVar(\graph^{k}_{t-1} | \graph_{t}) \right) \right]}_{L_{t-1}^{k}}$} \\
    &\resizebox{.55\linewidth}{!}{$\displaystyle L_0 = \sum_{k} \underbrace{\E_{q^k(\graph^{k}_{1}|\graph^{k}_0)} \left[ -\log p^k_\networkVar(\graph^{k}_{0} | \graph_{1}) \right]}_{L_0^{k}}.$}
    \end{align}
\end{subequations}
The joint loss, \ie Eq.~\eqref{eq:graph_variational_bound}, can be rearranged by domains:
\begin{equation}
    \resizebox{0.88\linewidth}{!}{$\displaystyle 
    L_{vb} = \sum_{t=1}^{T} L_{t-1} = \sum_{t=1}^{T} \left(\sum_{k} L^k_{t-1} \right) = \sum_k \left(\sum_{t=1}^{T} L^k_{t-1} \right).
    $}
    \label{eq:loss_by_domain}
\end{equation}
Each $\sum_{t=1}^{T} L^k_{t-1}$ recovers the standard variational diffusion loss for domain $k$, with the reverse model $p_\networkVar^k$ conditioned on the joint state $\graph_t$. 
This lets us reuse domain-specific implementations, \eg the \gls{ddpm} reparameterization~\cite{Ho20neurips-DiffusionModel} and the \gls{d3pm} state-transition parameterization~\cite{Austin21neurips-D3PM, Gu22cvpr-VQ-Diffusion}.
To adjust the relative importance of each domain, we apply weights $\lambda^k$ to each domain-specific loss (values in \Cref{app:sec:network}):
\begin{equation}
    L_{vb} = \sum_k \lambda^k \left(\sum_{t=1}^{T} L^k_{t-1} \right).
\end{equation}

For the room-level scene graph, $\graph = \graph^{\semanticVar} \times \graph^{\ContinuousVar} \times \graph^{\edgeVar}$, where $\graph^{\semanticVar} \in \discreteDomain^\numNodes$, $\graph^{\ContinuousVar} \in \Real{\nodeDim \times \numNodes}$, and $\graph^{\edgeVar} \in \{0,1\}^{\numNodes(\numNodes-1)/2}$.
In our experiments, we consider both a discrete (binary) parameterization of $\graph^{\edgeVar}$ and a continuous relaxation $\tilde{\graph}^{\edgeVar} \in [-1, 1]^{\numNodes(\numNodes-1)/2}$, mapping $\{0,1\}$ to $\{-1,+1\}$ and injecting Gaussian noise as in~\cite{Jo22icml-GDSS}.
For continuous relaxation of $\graph^{\semanticVar}$, we use the standard \gls{ddpm} formulation with one-hot encoding and Gaussian noise.
We consider all four combinations of discrete and continuous parameterizations for $(\semanticVar, \edgeVar)$.

\begin{figure}[t!]
    \centering
    \includegraphics[trim={20 400 335 20},clip,width=.95\linewidth]{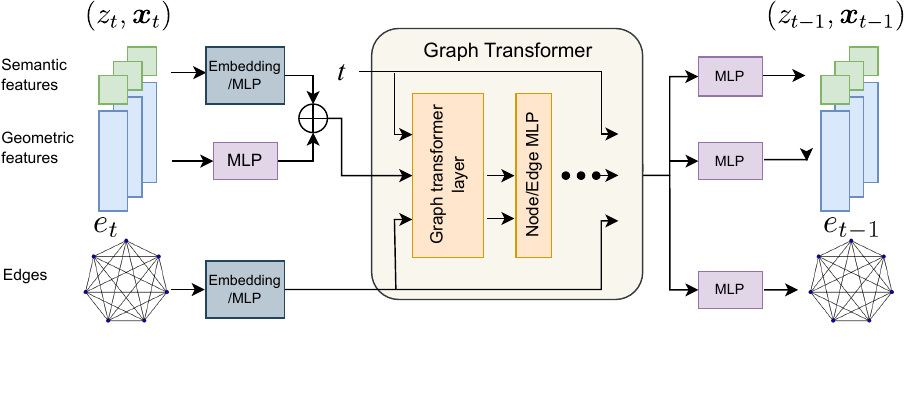} 
    \vspace{-20pt}
    \caption{Denoising architecture. Graph diffusion model over mixed feature domains.}
    \label{fig:architecture}
    \vspace{-15pt}
\end{figure}

\subsubsection{Denoising Network}
We show the denoising architecture in Fig.~\ref{fig:architecture} and include hyperparameters in \Cref{app:sec:network}.
The network takes $\graph_t =(\{( \semanticVar^{i}_t, \ContinuousVar^{i}_t)\}_{1 \leq i \leq \numNodes}, \{ \edgeVar^{ij}_t \}_{1 \leq i < j \leq \numNodes})$ at step $t$ and predicts the reverse-process parameters for each domain---either categorical distributions or Gaussian means (with fixed covariance), depending on the parameterization.
At test time, we sample these distributions to obtain $\graph_{t-1}$.
Inspired by latent diffusion models, the shape feature $\ShapeVar^i_t$ in $\ContinuousVar^i_t$ lives in the latent space of a 2D point cloud autoencoder. We denote the encoder as $f_{\cal{E}}: \Real{N_{in} \times 2} \rightarrow {\cal H}$ and decoder as $f_{\cal{D}}: {\cal H} \rightarrow \Real{N_{out} \times 2}$, where $N_{in}$ and $N_{out}$ are input and output point counts, respectively. We pre-train this autoencoder and regularize the latent space to $[-1, 1]$ for stability, \ie ${\cal H} = [-1, 1]^{\nodeDim-3}$. 

\myParagraph{Feature encoder}
The node input combines semantic label $\semanticVar^i_t$ and geometric feature $\ContinuousVar^i_t$. We encode $\ContinuousVar^i_t$ with an \gls{mlp}, and encode discrete features ($\semanticVar^i_t$ and $\edgeVar^{ij}_t$) either as one-hot vectors processed by an \gls{mlp} or as integers mapped to learned embeddings. We concatenate the semantic and geometric embeddings to form the node embedding.

\myParagraph{Graph transformer backbone}
The denoising backbone must capture all pairwise interactions while remaining permutation-invariant, for which we adopt the fully connected graph transformer from DiGress~\cite{Vignac23iclr-DiGress}.

\myParagraph{Feature extractor}
We use domain-specific \gls{mlp} feature extractors to predict reverse-process parameters: continuous features map to Gaussian means (fixed covariance), while discrete features map to categorical distributions. We evaluate all four combinations of discrete/continuous parameterizations for $(\semanticVar,\edgeVar)$ in the ablation study (\Cref{sec:ablation_study}).

\myParagraph{Shape post-processing}
The backward diffusion process predicts the latent shape feature $\ShapeVar_0^i\in {\cal H}$, which is decoded to a 2D point cloud via $f_{\cal{D}}$. We convert this point cloud into a polygonal boundary using a lightweight post-processing pipeline: we remove outlier points whose nearest-neighbor distance exceeds 0.5~m, order the remaining points by minimizing the polygon perimeter (TSP-style ordering), and simplify the boundary using the Ramer--Douglas--Peucker algorithm with a tolerance of 0.5~m.
On the 3D-FRONT~\cite{Fu21iccv-3dFront} dataset, we additionally adjust the polygon vertices to enforce axis-aligned consecutive edges while minimizing overall displacement, since the dataset contains only axis-aligned floor plans. Additional details are provided in \Cref{app:post_processing}.

\subsubsection{Partial Conditioning}
3DSG prediction is often conditioned on partial observations. In robotics, partial scene graphs result from incomplete exploration, and predicting the unexplored scene aids future planning. 
We consider two types of room conditions: (1) fully explored rooms, where the floor plan is known, \ie $\ContinuousVar^i_0=\ContinuousVar^{i,\known} = (\PosVar^{i,\known},\ShapeVar^{i,\known})$, and (2) partially explored rooms, where $\ContinuousVar^i_0$ must cover the conditional input boundary $\ContinuousVar^{i,\partialk} = (\PosVar^{i,\partialk},\ShapeVar^{i,\partialk})$. Known semantic labels $\semanticVar^{i,\known}$ and edge connectivity $\edgeVar^{ij,\known}$ (when both rooms are fully explored) are handled analogously.

We denote the conditioning graph for fully explored rooms by $\graph_0^{\known}$ and its feature mask by $m$, so $m \odot \graph_0^{\known}$ represents the known regions.\footnote{$\odot$ denotes element-wise (Hadamard) multiplication.}
For partially explored rooms, we denote the partial conditioning graph by $\graph_0^\partialk$ with mask $m^\prime$, so $m^\prime \odot \hat{\graph}_t^{\partialk}$ contains only $\ContinuousVar^{i, \partialk}$. $m^\prime$ and $m$ are mutually exclusive.
To handle partial constraints, we define $\rho(\cdot, \cdot | {\cal E}, {\cal D})$, parameterized by the point cloud autoencoder, to decode room boundaries, take their union in boundary space, and re-encode only the geometric feature $\ContinuousVar$; semantic labels and edges are fixed by masking. The denoising step is:
\begin{subequations}
\begin{align}
    \graph_t^{\unknown} &\sim p_\networkVar(\graph_t | \graph_{t+1}) \tag{\theequation a} \\
    \graph_t^{\known} &\sim q(\graph_t | \graph_0=\graph_0^{\known}) \tag{\theequation b} \\
    \hat{\graph}_t^{\partialk} &= \rho (\graph_t^{\unknown}, \graph_0^{\partialk}) \tag{\theequation c} \\
    \graph_t &= m \odot \graph_t^{\known} + m^\prime \odot \hat{\graph}_t^{\partialk} \notag \\
             &\quad + (1-m^\prime-m) \odot \graph_t^{\unknown} \tag{\theequation d}
\end{align}
\end{subequations}
This ensures $\graph_0$ satisfies both full and partial room constraints; in practice, we apply $\rho$ only late in reverse diffusion, when samples are less noisy.
When $m^\prime$ is empty, combining the corrupted known graph $\graph_t^{\known}$ and the predicted $\graph_t^{\unknown}$ at each step reduces to masking-based layout completion~\cite{Hu26wacv-midiffusion}. 
\subsection{Object-level Scene Graph Generation}
Similar to the room-level graph, the object-level graph in room $r$ can be denoted as $\graph^{O, r} = (\nodes^{O, r}, \edges^{O, r})$. 
We drop object edge synthesis since most inter-object relationships can be inferred from geometric placement, and focus on $\nodes^{O, r}$. We drop the room index $r$ for simplicity and assume object layouts are synthesized independently across rooms given their parent room nodes.
Each object node contains a semantic label $\semanticVar^{O,i}\in \discreteDomain$ and a geometric feature $\ContinuousVar^{O,i} \in \Real{\nodeDim}$.
With at most $M$ objects per room, object-level scene graph generation amounts to synthesizing $\graph^{O} \approx (\{( \semanticVar^{O,i}, \ContinuousVar^{O,i})\}_{1 \leq i \leq M})$.
This is a well-studied problem; existing approaches typically use position, bounding box, and vertical-axis rotation as the geometric feature, and retrieve furniture models to render the layout. We use MiDiffusion~\cite{Hu26wacv-midiffusion} for object layout synthesis, as it uses a similar masking strategy to our room-level approach, avoiding the need for retraining, and demonstrates superior performance compared to discrete or continuous diffusion models and autoregressive baselines.

%% file: tex/experiment.tex
\section{Experiment}\label{sec:experiment}
We conduct ablation studies comparing discrete and continuous choices for node and edge label formulations in our room-layer completion module. We then compare our best model, without retraining, against state-of-the-art methods on floor plan completion tasks using both training-like inputs and realistic partial scene graphs from a different dataset. Finally, we demonstrate integration of our hierarchical scene graph prediction pipeline with Hydra~\cite{Hughes24ijrr-hydraFoundations}, a real-time scene graph construction pipeline, and MiDiffusion~\cite{Hu26wacv-midiffusion}.

\myParagraph{Datasets}
We use the 3D-FRONT~\cite{Fu21iccv-3dFront} benchmark, a synthetic apartment dataset created by indoor designers. 
We determine connectivity by identifying doors or large openings between rooms, and filter out scenes with non-traversable rooms. We also limit the maximum length of any room to 12\,m. After these two filtering steps, we obtain a total of 5,107 scenes, which is 75\% of the original data, and split them into 3,572, 511, and 1,024 scenes for training, validation, and testing, respectively. These scenes contain a total of 36,284 rooms, and we map them to 17 room categories. We use rotation augmentations in $90^\circ$ increments during training as in~\cite{Tang24cvpr-DiffuScene} to be consistent with the baselines.

For out-of-distribution evaluation, we also use partial scene graphs constructed by Hydra~\cite{Hughes24ijrr-hydraFoundations} from 6 single-floor apartment scenes in the \gls{mp3d} dataset~\cite{Chang173dv-Matterport3D}, using sequences of RGB-D images, poses, and object-level semantic segmentation. We sample 5 partial scene graphs from each scene, for a total of 30 partial scene graphs that closely mimic real-world inputs.
Finally, we collect two real-world apartment datasets using an Intel RealSense D455 camera for scanning and an Intel RealSense T265 camera for odometry, and construct hierarchical scene graphs with Hydra for the real-world pipeline demonstration.

\myParagraph{Baselines}
We compare our room-level scene graph generation module with state-of-the-art occupancy-based and \gls{llm}-based methods.
For voxel-based \gls{ssc}, we include Lee~\textit{et al.}~\cite{Lee23arxiv-diffusionSSC}, which generates scenes conditioned on voxel occupancies using a discrete diffusion model (denoted SSC).
We also compare with AnyHome~\cite{Fu24eccv-AnyHome} and Holodeck~\cite{Yang24cvpr-Holodeck}, two \gls{llm}-based methods that generate floor plans from text descriptions. AnyHome first predicts room connectivity and then passes it to HouseGAN++~\cite{Nauata21cvpr-House-GAN++} to generate the final layout; Holodeck directly predicts room-corner coordinates.

\myParagraph{Implementation}
We generate 1,024 scenes per model in each experiment. For the unconditioned setting, we sample layouts without conditioning inputs. For scene completion, we condition generation on partial observations. For 3D-FRONT, partial scene graphs are obtained by restricting to randomly sampled rectangular patches, while \gls{mp3d} partial scene graphs are constructed by Hydra.
For fair comparisons, we use publicly available implementations for all baseline methods and provide all approaches with the same inputs during testing.

\myParagraph{Evaluation Metrics}
We evaluate room-graph prediction in four aspects.
(1) \textbf{Layout realism}: FID/KID between rendered top-down projections similar to~\cite{Hu26wacv-midiffusion} (256$\times$256 over 12\,m$\times$12\,m).
(2) \textbf{Graph structure}: MMD over degree and clustering statistics following~\cite{You18icml-GraphRNNGenerating,Jo22icml-GDSS,Vignac23iclr-DiGress}.
(3) \textbf{Geometry}: mean pairwise room overlap (IoU\%) and mean distance between connected rooms.
(4) \textbf{Labels/edges}: room-label KL, edge-matrix L1/L2 (pairwise connectivity probabilities over room pairs), and room-count KL.

\subsection{Ablation Study}\label{sec:ablation_study}
We compare discrete vs. continuous formulations for node labels and edges in our room-level scene graph generation module. Using the 3D-FRONT training set, we train four variants covering all combinations (Disc./Cont. denote discrete/continuous; L/E denote labels/edges in Table~\ref{tab:unconditioned}).
We evaluate in the unconditioned setting, where each model synthesizes room graphs without conditioning inputs. Since the models are trained to match the 3D-FRONT training distribution, we report all metrics by comparing generated layouts against the training set. As reference, the last row reports metrics computed between test and training sets, providing an oracle baseline for same-distribution performance.
Discrete edge representation performs better on almost all metrics. Prior work on graph generation~\cite{Jo22icml-GDSS,Vignac23iclr-DiGress} suggests discrete representations require fewer diffusion steps (\ie are easier to learn). In our joint setting, discrete edges converge faster, improving connectivity prediction and dominating over continuous quantities at inference. Intuitively, discrete edges also better match the binary nature of room connectivity.
For node labels, discrete labels slightly improve room-count and label-distribution metrics, while continuous node labels perform better on most others. This suggests a trade-off between geometry (which also affects layout realism) and label prediction; keeping labels continuous (typically slower to converge) improves overall performance.
Overall, the model with continuous labels and discrete edges (``Cont. L + Disc. E'') provides the best overall trade-off across metrics, so we use it in subsequent experiments.

\input{tabs/unconditioned}
\input{figures/qualitative_combined}

\subsection{Scene Completion}\label{sec:scene_completion}
We compare our best model (from the ablation study ``Cont. L + Disc. E'') against baseline methods on floor plan completion tasks. Our model handles partial inputs in a zero-shot manner without retraining. For the SSC baseline, we use randomly sampled patches as well as full layouts from the 3D-FRONT training set.
For the \gls{llm}-based methods (\ie AnyHome and Holodeck), which are not originally designed for partial-graph completion, we adapt the same partial evidence to their text-conditioned interfaces by converting input scene graphs to text instructions. For AnyHome, we additionally provide label masks as detailed geometric context for the input.
This preserves each method's native interface while aligning the observed evidence across baselines.
We report comparisons under two setups: (1) in-distribution inputs from the 3D-FRONT test set, and (2) out-of-distribution partial scene graphs from \gls{mp3d}.
Qualitative comparisons are shown in Fig.~\ref{fig:conditioned}, with additional examples in \Cref{app:additionalImages}.

\newpage
\input{tabs/conditioned}
\input{tabs/conditioned_mp3d}

\subsubsection{In-distribution 3D-FRONT partial scene graphs\label{sec:conditioned_3dfront}}
We first evaluate our model on in-distribution partial scene graphs from the 3D-FRONT test set. For each test scene, we obtain a partial scene graph by sampling a rectangular patch, matching the training setup. For each method, we generate 1,024 layouts and compare them with 1,024 reference layouts from the test set. In the GT row, set-comparison metrics are identical-set comparisons and reported as ``--'', while single-set metrics are still computed on GT layouts. The two \gls{llm}-based approaches fail to produce outputs for some cases (\ie they cannot generate results that satisfy the language constraints): AnyHome generates 504 scenes and Holodeck 1,019 scenes out of 1,024.
The first three columns of Fig.~\ref{fig:conditioned} show qualitative comparisons across methods.
Both AnyHome and Holodeck struggle to satisfy geometric constraints. Moreover, AnyHome uses a backbone trained on a different dataset, leading to layouts that deviate from the 3D-FRONT distribution. Holodeck's reliance on axis-aligned rectangles limits the geometric diversity and realism of generated environments in settings that require complex room shapes.
SSC can yield predictions that are inconsistent with the input labels. 
Our approach produces layouts that are closer to the ground truth and better capture complex room shapes, but it is less precise than the pixel-based SSC baseline at local room-boundary alignment. 
Quantitative results are reported in Table~\ref{tab:conditioned}. The domain-specific learning methods (SSC and ours) achieve substantially better layout realism than the \gls{llm}-based approaches, which struggle with geometric prediction. SSC cannot predict room connectivity, so we report the corresponding metrics as ``n/a''. Consistent with the qualitative results, our main failure mode is slight room overlap near room boundaries, whereas the baselines avoid overlap by design. Overall, our method outperforms all baselines on nearly all other metrics.

\subsubsection{Out-of-distribution \gls{mp3d} partial scene graphs\label{sec:conditioned_mp3d}}
To compare SSC and ours on out-of-distribution partial scene graphs, we generate 1,024 scenes conditioned on inputs from 30 partial scene graphs. These partial inputs are constructed by Hydra~\cite{Hughes24ijrr-hydraFoundations} from 6 single-floor apartment scenes in the \gls{mp3d} dataset~\cite{Chang173dv-Matterport3D}.
\Cref{app:hydra_processing} includes details on how we extract partial room geometry from Hydra outputs. 
Results are summarized in Table~\ref{tab:conditioned_mp3d}. As expected, most metrics degrade relative to Table~\ref{tab:conditioned} due to the 3D-FRONT--\gls{mp3d} domain gap. The same GT-row convention for identical-set comparisons applies, and the reference set contains the 6 available \gls{mp3d} ground-truth scenes.
Our method substantially outperforms SSC across all metrics except IoU, where SSC achieves zero overlap by design, suggesting promising generalization to out-of-distribution inputs.
We include AnyHome (334 successful generations) and Holodeck (989 successful generations) for completeness. The \gls{llm}-based methods again struggle with geometric prediction, consistent with the qualitative results in the last three columns of Fig.~\ref{fig:conditioned}.

\subsection{Integration with Hydra and MiDiffusion}\label{sec:integration}
Finally, we combine our hierarchical scene graph completion pipeline with Hydra~\cite{Hughes24ijrr-hydraFoundations} and MiDiffusion~\cite{Hu26wacv-midiffusion} to enable completion from real robot-collected data.
We use a top-down formulation in which each child layer is generated conditioned on parent geometry and available observations.
Hydra constructs multi-layer \glspl{3dsg} from robot exploration. For indoor scenes, we use Hydra's Place layer to extract partial room geometry and connectivity as inputs to our room-level completion module, then apply MiDiffusion to synthesize object layouts in each completed room.
Details on partial-evidence extraction and fusion of Hydra outputs into a \gls{bro} hierarchy are provided in \Cref{app:hydra_processing}.

\Cref{fig:apartment_integration} summarizes a real-world completion example: partial evidence with camera trajectory (\subref{fig:apartment_partial}), the completed floor plan (\subref{fig:apartment_sc}), and the resulting layered scene graph (\subref{fig:apartment_sc_3d}). Additional visualizations are provided in the video attachment.

%% file: tabs/unconditioned.tex
\begin{table*}[!th]
\caption{Ablation: unconditioned room-level scene graph generation. Bold indicates entries closest to the GT test row, and underlines indicate the second closest.\label{tab:unconditioned}}
\vspace{-5pt}
\resizebox{\textwidth}{!}{%
\begin{tabular}{lcccccccccc}
\toprule
\multirow{2}{*}{Approach} & \multicolumn{2}{c}{Layout}                                   & \multicolumn{2}{c}{Graph MMD}  & \multicolumn{2}{c}{Geometry}         & \multicolumn{3}{c}{Label}                                                                                    & \multicolumn{1}{c}{Num Nodes}                                                                     \\
\cmidrule(lr){2-3} \cmidrule(lr){4-5} \cmidrule(lr){6-7} \cmidrule(lr){8-10} \cmidrule(lr){11-11}
                         & \multicolumn{1}{c}{FID $\downarrow$} & \multicolumn{1}{c}{KID x 0.001 $\downarrow$}  & \multicolumn{1}{c}{Degree x 0.01 $\downarrow$} & \multicolumn{1}{c}{Clustering x 0.01 $\downarrow$} & \multicolumn{1}{c}{IoU $\downarrow$} & \multicolumn{1}{c}{Dist - connected} & \multicolumn{1}{c}{Node KL x 0.01 $\downarrow$} & \multicolumn{1}{c}{Edge diff (L1) $\downarrow$} & \multicolumn{1}{c}{Edge diff (L2) $\downarrow$} & \multicolumn{1}{c}{KL x 0.01 $\downarrow$}   \\ 
\midrule
Disc.\ L + disc.\ E & \underline{23.39} & \underline{20.11} & 0.093 & \textbf{0.023} & \underline{0.016} & \underline{0.71} & \underline{0.153} & \underline{0.118} & \underline{0.063} & \textbf{1.95} \\
Disc.\ L + cont.\ E & 26.92 & 24.74 & 0.134 & 0.032 & 0.020 & \underline{0.79} & \textbf{0.160} & 0.303 & 0.138 & \underline{2.15} \\
Cont.\ L + disc.\ E & \textbf{20.95} & \textbf{17.16} & \textbf{0.028} & \underline{0.024} & \textbf{0.009} & \textbf{0.75} & 0.272 & \textbf{0.142} & \textbf{0.065} & 2.79 \\
Cont.\ L + cont.\ E & 24.48 & 21.64 & \underline{0.080} & 0.051 & 0.020 & \textbf{0.75} & 1.349 & 0.291 & 0.134 & 3.64 \\
\midrule
GT test & 4.06 & 0.04 & 0.020 & 0.004 & 0.000 & 0.75 & 0.191 & 0.170 & 0.085 & 0.98 \\
\bottomrule
\end{tabular}%
}
\end{table*}

%% file: figures/qualitative_combined.tex

\begin{figure*}[!ht]
    \centering

    \centering

    \newcommand{\qualframedimg}[1]{%
        \fbox{\includegraphics[width=\dimexpr\linewidth-2\fboxsep-2\fboxrule\relax]{#1}}%
    }

     \begin{minipage}[t]{0.09\linewidth}
       {\footnotesize AnyHome\\~\cite{Fu24eccv-AnyHome}}
    \end{minipage}%
    \begin{minipage}{0.15\linewidth}
        \centering
        \qualframedimg{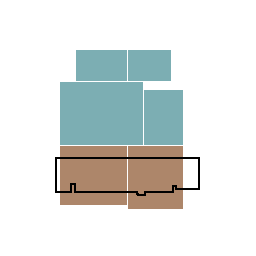}%
    \end{minipage}%
    \begin{minipage}{0.15\linewidth}
        \centering
        \qualframedimg{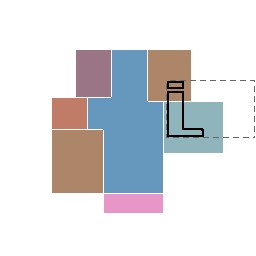}%
    \end{minipage}%
    \begin{minipage}{0.15\linewidth}
        \centering
        \qualframedimg{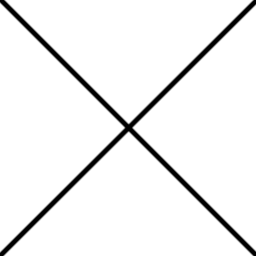}%
    \end{minipage}%
    \hspace{0.01\linewidth}%
    \begin{minipage}{0.15\linewidth}
        \centering
        \qualframedimg{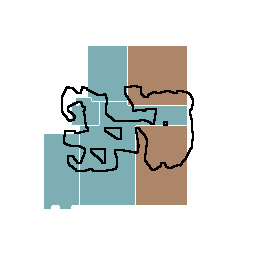}%
    \end{minipage}%
    \begin{minipage}{0.15\linewidth}
        \centering
        \qualframedimg{figures/cross_256x256.png}%
    \end{minipage}%
    \begin{minipage}{0.15\linewidth}
        \centering
        \qualframedimg{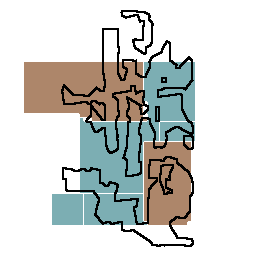}%
    \end{minipage}%

     \begin{minipage}[t]{0.09\linewidth}
       {\footnotesize Holodeck\\~\cite{Yang24cvpr-Holodeck}}
    \end{minipage}%
    \begin{minipage}{0.15\linewidth}
        \centering
        \qualframedimg{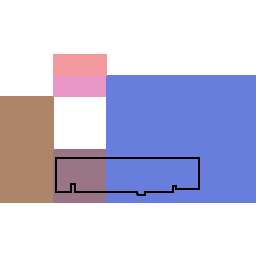}%
    \end{minipage}%
    \begin{minipage}{0.15\linewidth}
        \centering
        \qualframedimg{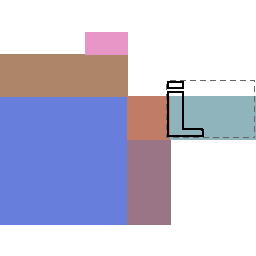}%
    \end{minipage}%
    \begin{minipage}{0.15\linewidth}
        \centering
        \qualframedimg{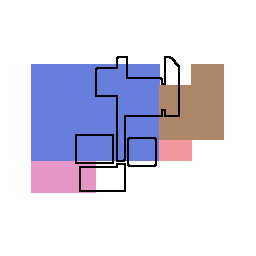}%
    \end{minipage}%
    \hspace{0.01\linewidth}%
    \begin{minipage}{0.15\linewidth}
        \centering
        \qualframedimg{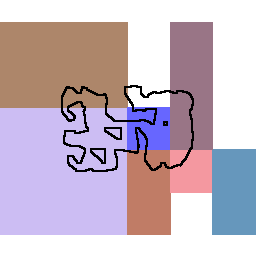}%
    \end{minipage}%
    \begin{minipage}{0.15\linewidth}
        \centering
        \qualframedimg{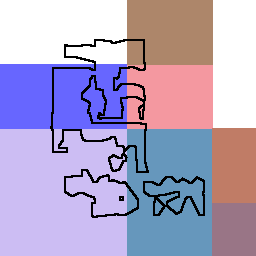}%
    \end{minipage}%
    \begin{minipage}{0.15\linewidth}
        \centering
        \qualframedimg{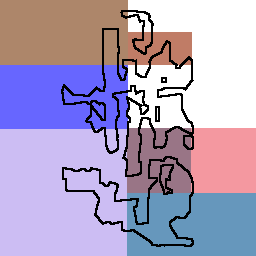}%
    \end{minipage}%

     \begin{minipage}[t]{0.09\linewidth}
       {\footnotesize SSC\\~\cite{Lee23arxiv-diffusionSSC}}
    \end{minipage}%
    \begin{minipage}{0.15\linewidth}
        \centering
        \qualframedimg{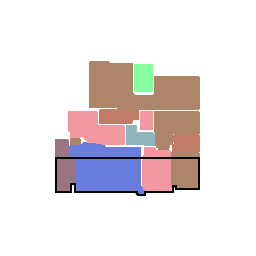}%
    \end{minipage}%
    \begin{minipage}{0.15\linewidth}
        \centering
        \qualframedimg{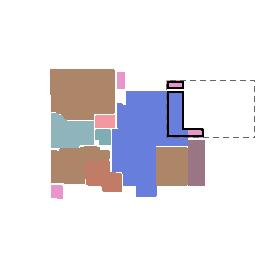}%
    \end{minipage}%
    \begin{minipage}{0.15\linewidth}
        \centering
        \qualframedimg{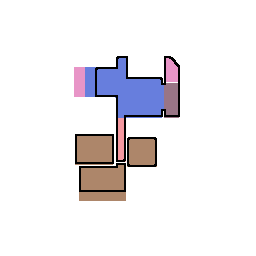}%
    \end{minipage}%
    \hspace{0.01\linewidth}%
    \begin{minipage}{0.15\linewidth}
        \centering
        \qualframedimg{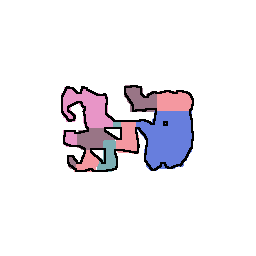}%
    \end{minipage}%
    \begin{minipage}{0.15\linewidth}
        \centering
        \qualframedimg{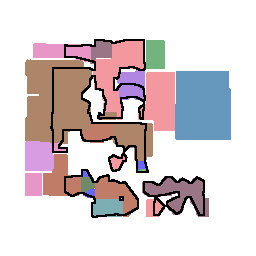}%
    \end{minipage}%
    \begin{minipage}{0.15\linewidth}
        \centering
        \qualframedimg{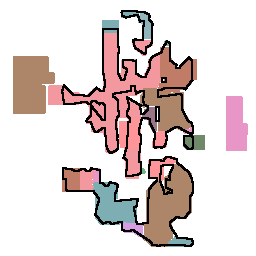}%
    \end{minipage}%

     \begin{minipage}[t]{0.09\linewidth}
       {\footnotesize Ours}
    \end{minipage}%
    \begin{minipage}{0.15\linewidth}
        \centering
        \qualframedimg{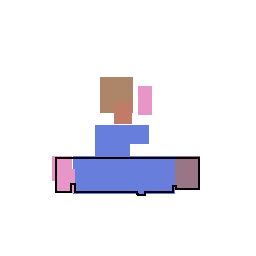}%
    \end{minipage}%
    \begin{minipage}{0.15\linewidth}
        \centering
        \qualframedimg{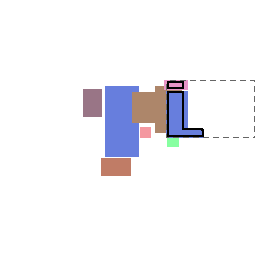}%
    \end{minipage}%
    \begin{minipage}{0.15\linewidth}
        \centering
        \qualframedimg{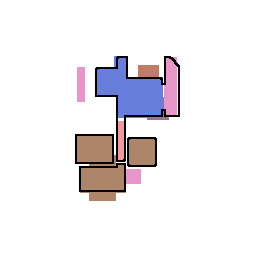}%
    \end{minipage}%
    \hspace{0.01\linewidth}%
    \begin{minipage}{0.15\linewidth}
        \centering
        \qualframedimg{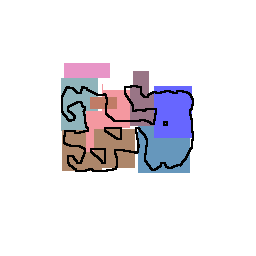}%
    \end{minipage}%
    \begin{minipage}{0.15\linewidth}
        \centering
        \qualframedimg{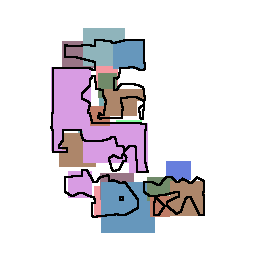}%
    \end{minipage}%
    \begin{minipage}{0.15\linewidth}
        \centering
        \qualframedimg{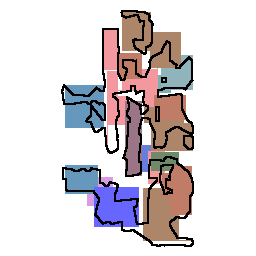}%
    \end{minipage}%

     \begin{minipage}[t]{0.09\linewidth}
       {\footnotesize Ground Truth}
    \end{minipage}%
    \begin{minipage}{0.15\linewidth}
        \centering
        \qualframedimg{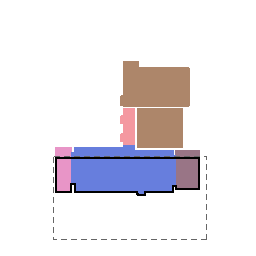}%
    \end{minipage}%
    \begin{minipage}{0.15\linewidth}
        \centering
        \qualframedimg{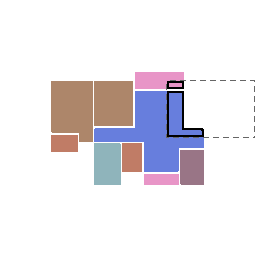}%
    \end{minipage}%
    \begin{minipage}{0.15\linewidth}
        \centering
        \qualframedimg{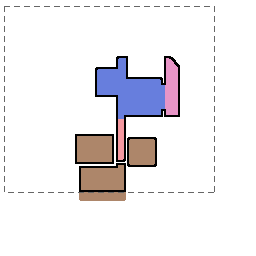}%
    \end{minipage}%
    \hspace{0.01\linewidth}%
    \begin{minipage}{0.15\linewidth}
        \centering
        \qualframedimg{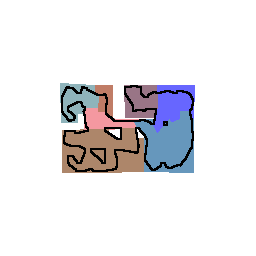}%
    \end{minipage}%
    \begin{minipage}{0.15\linewidth}
        \centering
        \qualframedimg{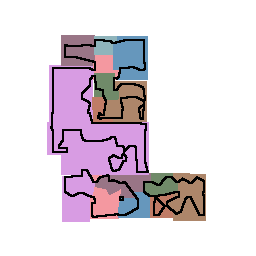}%
    \end{minipage}%
    \begin{minipage}{0.15\linewidth}
        \centering
        \qualframedimg{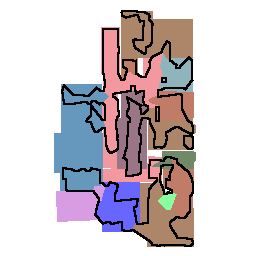}%
    \end{minipage}%

    \begin{minipage}{\linewidth}
        \centering
        \includegraphics[width=\linewidth,trim=0 0 0 4pt,clip]{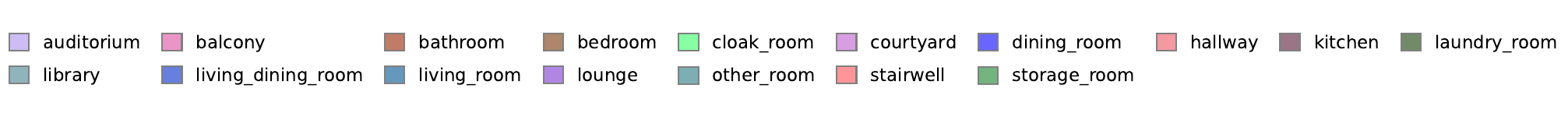}
    \end{minipage}

    \vspace{-10pt}
    \caption{Scene completion examples (left: 3D-FRONT; right: MP3D). Room colors are rendered at 60\% opacity to contrast black input boundaries. Conditional partial-scene boundaries are highlighted in black, and conditioned regions are expected to match ground-truth regions. SSC and our model are trained on 3D-FRONT partial inputs obtained by randomly sampling rectangular patches (dashed lines), and tested on both 3D-FRONT (in-distribution) and MP3D (out-of-distribution). \label{fig:conditioned}}
    \vspace{-12pt}
\end{figure*}

%% file: tabs/conditioned.tex

\begin{table*}[!t]
\caption{Conditioned room-level scene graph generation given sampled patches from the 3D-FRONT test set. Notation: ``--'' denotes identical-set comparisons and ``n/a'' denotes unavailable metrics. Bold/underline indicate best/second-best values (closest to the GT row when available, otherwise lowest among compared approaches).\label{tab:conditioned}}
\vspace{-5pt}
\resizebox{\textwidth}{!}{%
\begin{tabular}{lcccccccccc}
\toprule
\multirow{2}{*}{Approach} & \multicolumn{2}{c}{Layout}                                   & \multicolumn{2}{c}{Graph MMD}  & \multicolumn{2}{c}{Geometry}         & \multicolumn{3}{c}{Label}                                                                                    & \multicolumn{1}{c}{Num Nodes}                                                                     \\
\cmidrule(lr){2-3} \cmidrule(lr){4-5} \cmidrule(lr){6-7} \cmidrule(lr){8-10} \cmidrule(lr){11-11}
                         & \multicolumn{1}{c}{FID $\downarrow$} & \multicolumn{1}{c}{KID x 0.001 $\downarrow$}  & \multicolumn{1}{c}{Degree x 0.01 $\downarrow$} & \multicolumn{1}{c}{Clustering x 0.01 $\downarrow$} & \multicolumn{1}{c}{IoU $\downarrow$} & \multicolumn{1}{c}{Dist - connected} & \multicolumn{1}{c}{Node KL x 0.01 $\downarrow$} & \multicolumn{1}{c}{Edge diff (L1) $\downarrow$} & \multicolumn{1}{c}{Edge diff (L2) $\downarrow$} & \multicolumn{1}{c}{KL x 0.01 $\downarrow$}   \\ 
\midrule
AnyHome~\cite{Fu24eccv-AnyHome}  & 124.13 & 119.10 & \textbf{1.471} & 0.156 & \textbf{0.000} & \textbf{0.745} & 177.030 & 3.411 & 2.096 & 138.85 \\
Holodeck~\cite{Yang24cvpr-Holodeck} & 123.57 & 162.92 & 5.845 & \underline{0.098} & \underline{0.001} & 1.131 & \underline{1.810} & \underline{1.643} & \underline{0.905} & 64.75 \\
\midrule
SSC~\cite{Lee23arxiv-diffusionSSC}      & \underline{30.79} & \underline{30.79} & n/a & n/a & \textbf{0.000} & n/a & 5.951 & n/a & n/a & \textbf{9.86} \\
Ours     & \textbf{26.33} & \textbf{23.80} & \underline{2.677} & \textbf{0.089} & 0.019 & \underline{0.705} & \textbf{0.423} & \textbf{0.940} & \textbf{0.509} & \underline{37.65} \\
\midrule
GT test  & -- & -- & -- & -- & 0.000 & 0.750 & -- & -- & -- & -- \\
\bottomrule
\end{tabular}%
}
\vspace{-3pt}
\end{table*}

%% file: tabs/conditioned_mp3d.tex

\begin{table*}[!t]
\caption{Conditional room-level scene graph generation given partial MP3D scene graphs. Notation and highlighting follow Table~\ref{tab:conditioned}.\label{tab:conditioned_mp3d}}
\vspace{-5pt}
\resizebox{\textwidth}{!}{%
\begin{tabular}{lcccccccccc}
\toprule
\multirow{2}{*}{Approach} & \multicolumn{2}{c}{Layout} & \multicolumn{2}{c}{Graph MMD} & \multicolumn{2}{c}{Geometry} & \multicolumn{3}{c}{Label} & \multicolumn{1}{c}{Num Nodes}\\
\cmidrule(lr){2-3} \cmidrule(lr){4-5} \cmidrule(lr){6-7} \cmidrule(lr){8-10} \cmidrule(lr){11-11}
                         & \multicolumn{1}{c}{FID $\downarrow$} & \multicolumn{1}{c}{KID x 0.001 $\downarrow$}  & \multicolumn{1}{c}{Degree x 0.01 $\downarrow$} & \multicolumn{1}{c}{Clustering x 0.01 $\downarrow$} & \multicolumn{1}{c}{IoU $\downarrow$} & \multicolumn{1}{c}{Dist - connected} & \multicolumn{1}{c}{Node KL x 0.01 $\downarrow$} & \multicolumn{1}{c}{Edge diff (L1) $\downarrow$} & \multicolumn{1}{c}{Edge diff (L2) $\downarrow$} & \multicolumn{1}{c}{KL x 0.01 $\downarrow$}   \\ 
\midrule
AnyHome~\cite{Fu24eccv-AnyHome}   & \underline{167.67} & 128.01 & 13.057 & 39.711 & \textbf{0.000} & \underline{0.781} & 376.82 & 9.667 & 4.696 & 635.138 \\
Holodeck~\cite{Yang24cvpr-Holodeck}   & 203.98 & 210.15 & \underline{6.659} & \underline{36.173} & \underline{0.002} & 1.051 & \textbf{2.53} & \underline{6.183} & \underline{2.630} & \underline{470.965} \\
\midrule
SSC~\cite{Lee23arxiv-diffusionSSC}  & 190.91 & \underline{106.66} & n/a & n/a & \textbf{0.000} & n/a & 72.59 & n/a & n/a & 972.150 \\
Ours      & \textbf{99.29} & \textbf{16.42} & \textbf{1.900} & \textbf{1.491} & 0.019 & \textbf{0.684} & \underline{13.139} & \textbf{1.403} & \textbf{0.627} & \textbf{126.541} \\
\midrule
GT MP3D   & -- & -- & -- & -- & 0.000 & 0.273 & -- & -- & -- & -- \\
\bottomrule
\end{tabular}%
}
\vspace{-15pt}
\end{table*}

%% file: tex/conclusion.tex
\section{Conclusion}\label{sec:conclusion}
We present a top-down framework for synthesizing hierarchical \glspl{3dsg}, focusing on room-level scene graph prediction.
At the room level, our mixed-domain graph diffusion model jointly predicts room semantics, boundaries, and inter-room connectivity, while supporting partial constraints through corruption and masking without retraining.
Ablations in unconditioned room-graph generation validate the benefits of discrete--continuous joint modeling.
In the conditioned setting, on both in-distribution (3D-FRONT) and out-of-distribution (\gls{mp3d}) partial scene graphs, our method improves layout realism and graph fidelity over occupancy-based and \gls{llm}-based baselines while better capturing complex room shapes.
Finally, we demonstrate integration with Hydra and MiDiffusion for real-world robotic scene completion.
This pipeline opens directions for joint room-object evaluation and downstream tasks such as exploration and object search.

%% file: tex/supplementary/implementation.tex
%

\subsection{Network and Implementation Details \label{app:implementation}}
We provide additional details on the network architecture, training hyper-parameters, and implementation for our room layout prediction module. 
Based on 3D-FRONT training data, we set the maximum number of rooms to $\numNodes=22$ and the maximum number of objects per room to $M=21$ with ``empty'' labels for padding purposes.

\subsubsection{Network Hyper-parameters \label{app:sec:network}}
The backbone of our architecture (\Cref{fig:architecture}) is a series of 8 graph-transformer blocks adopted from~\cite{Vignac23iclr-DiGress}. In the multi-head attention layer, the node dimension is set to 256 with 8 attention head, the edge and time dimensions are set to 8 and 64 respectively. In the feed-forward layer, the dimensions of node, edge, and time are set to 256, 8, and 64 respectively. 
Before passing to the graph transformer backbone, discrete features are embedded in a learned vector and continuous features are mapped by a 2-layer \gls{mlp}, both to feature-specific dimensions as in the graph transformer. For continuous feature representations, we set the hidden dimension to 256 for node features and to 8 for edge features.
The time embedding (\ie diffusion step) is computed using sinusoidal encoding.
For continuous features, output feature extractors are 2-layer \glspl{mlp} with hidden dimensions set to 256, 8, 16 respectively for node, edge, and time features. For discrete features, the output feature extractor is a 1-layer \gls{mlp} that produces a categorical distribution over respective dimensions.
For the domain loss weights, we use $\lambda^{\ContinuousVar}=1$ and $\lambda^{\semanticVar}=0.25$ in all experiments. For edge prediction, we set $\lambda^{\edgeVar}=1$ for continuous parameterization and $\lambda^{\edgeVar}=10$ for discrete parameterization.

\subsubsection{Implementation}
We use a learning rate of $l_r=2e^{-4}$ with $0.5$ decay every $10k$ epochs for $60k$ epochs using the Adam optimizer, a dropout ratio of 0.1 for multi-head attention and feed-forward layers in the transformer blocks. 
We use standard noise schedule from discrete and continuous domains, same as MiDiffusion~\cite{Hu26wacv-midiffusion}. 
We train all our models on a single NVIDIA V100 GPU with under 8GB of GPU RAM usage.
For a batch size of 256, the training time is around 70 hours.
At inference time, for partial room constraints we use position-only conditioning during most of the reverse process and apply the partial masking operator only in the last 10 denoising steps, when the samples are less noisy.


%% file: tex/supplementary/post_processing.tex
%

\subsection{Room Layout Processing from Decoded Point Clouds \label{app:post_processing}}

Our room-level diffusion model decodes each room boundary as a noisy 2D point cloud. For evaluation against polygon-based baselines, we convert the decoded points into a clean, closed polygon that (i) follows the decoded boundary, and (ii) enforces orthogonal (Manhattan-style) edges while preserving the vertex order.
Note that this computation is not necessary for practical applications if 2D point cloud boundary prediction is sufficient. We only perform this post-processing for evaluation purposes to ensure a fair comparison to polygon-based baselines.

Let $\mathcal{P}=\{\mathbf{p}_j\}_{j=1}^{N}$ denote the decoded 2D boundary points in metric coordinates. The full post-processing pipeline is:
\begin{enumerate}
	\item \textbf{Outlier removal.} Remove isolated points whose nearest-neighbor distance exceeds $0.5$~m.
	\item \textbf{Optimization 1: boundary ordering.} Recover a cyclic vertex ordering by (approximately) solving a \gls{tsp} on the remaining points.
	\item \textbf{Boundary simplification.} Apply the Ramer--Douglas--Peucker algorithm to the ordered polyline with tolerance $0.5$~m to obtain a simplified polygonal boundary.
	\item \textbf{Optimization 2 (3D-FRONT only): boundary denoising.} Project the simplified polygon to an axis-aligned one by a least-squares fit with Manhattan-edge constraints.
\end{enumerate}
We highlight the two optimization problems (Optimizations 1 and 2) below.

\myParagraph{Outlier removal}
We filter $\mathcal{P}$ by removing points whose distance to their nearest neighbor exceeds $\tau_{\mathrm{out}}$.
We denote the filtered point set by $\mathcal{P}'$.

\myParagraph{Optimization 1: boundary ordering}
The decoded points are unordered. We recover an approximate cyclic boundary ordering by solving a \gls{tsp} on the complete graph over $\mathcal{P}'$ with Euclidean edge weights,
\[
	w_{ij}=\lVert \mathbf{p}_i-\mathbf{p}_j\rVert_2.
\]
Equivalently, we seek a cyclic permutation $\sigma$ that (approximately) minimizes the perimeter length,
\[
    \min_{\sigma}\; \sum_{i=1}^{N'} \lVert \mathbf{p}_{\sigma(i+1)}-\mathbf{p}_{\sigma(i)}\rVert_2,\qquad \mathbf{p}_{\sigma(N'+1)}\equiv \mathbf{p}_{\sigma(1)},
\]
where $N'=|\mathcal{P}'|$.
The resulting tour $\sigma$ induces a cyclic ordering $\mathbf{p}_{\sigma(1)},\dots,\mathbf{p}_{\sigma(N')}$ that approximately follows the room perimeter.

\myParagraph{Boundary simplification with Ramer--Douglas--Peucker}
Given the ordered cycle, we apply the Ramer--Douglas--Peucker algorithm (tolerance $\tau_{\mathrm{rdp}}$) to obtain a simplified, ordered vertex sequence $\widetilde{\mathcal{P}}=\{\tilde{\mathbf{p}}_i\}_{i=1}^{M}$ (with $M\le N'$) that removes small zig-zag artifacts while preserving the overall shape.

\myParagraph{Optimization 2 (3D-FRONT only): boundary denoising via least-squares axis-aligned projection}
For 3D-FRONT, room boundaries are axis-aligned by construction. Given the simplified ordered vertices $\widetilde{\mathcal{P}}$, we solve for a cleaned set of points $\mathcal{Q}=\{\mathbf{q}_i\}_{i=1}^{M}$ with a one-to-one correspondence to $\widetilde{\mathcal{P}}$ (same order and cardinality). The goal is to stay close to the decoded boundary while enforcing that each consecutive segment is axis-aligned:
\begin{align}
\min_{\mathcal{Q}}\quad & \sum_{i=1}^{M}\lVert \mathbf{q}_i-\tilde{\mathbf{p}}_i\rVert_2^2 \\
\text{s.t.}\quad & (\mathbf{q}_{i+1}-\mathbf{q}_i)_x = 0,\\
& \qquad\text{or}\qquad (\mathbf{q}_{i+1}-\mathbf{q}_i)_y = 0,\quad i=1,\dots,M,\\
& \mathbf{q}_{M+1}\equiv\mathbf{q}_1.
\label{eq:hierdiff:axis_aligned_ls}
\end{align}

The ``or'' constraint can be enforced exactly via a \gls{miqp}: we introduce a binary variable $b_i\in\{0,1\}$ per edge, where $b_i=1$ selects a horizontal edge (enforcing $y_{i+1}=y_i$) and $b_i=0$ selects a vertical edge (enforcing $x_{i+1}=x_i$).
In practice, \glspl{miqp} can be solved efficiently using a big-$M$ relaxation.

%% file: tex/supplementary/hydra_processing.tex
%

\subsection{Partial Scene Graph Extraction from Hydra \label{app:hydra_processing}}

This appendix describes how we extract partial room-level observations from Hydra~\cite{Hughes24ijrr-hydraFoundations} and incorporate them into a building--room--object hierarchical scene graph for the out-of-distribution \gls{mp3d} experiment in \Cref{sec:scene_completion} and the full completion pipeline in \Cref{sec:integration}.

\begin{figure}[ht]
    \centering
    \begin{subfigure}[t]{0.56\linewidth}
        \centering
        \includegraphics[width=\linewidth,trim=0 0 0 10mm,clip]{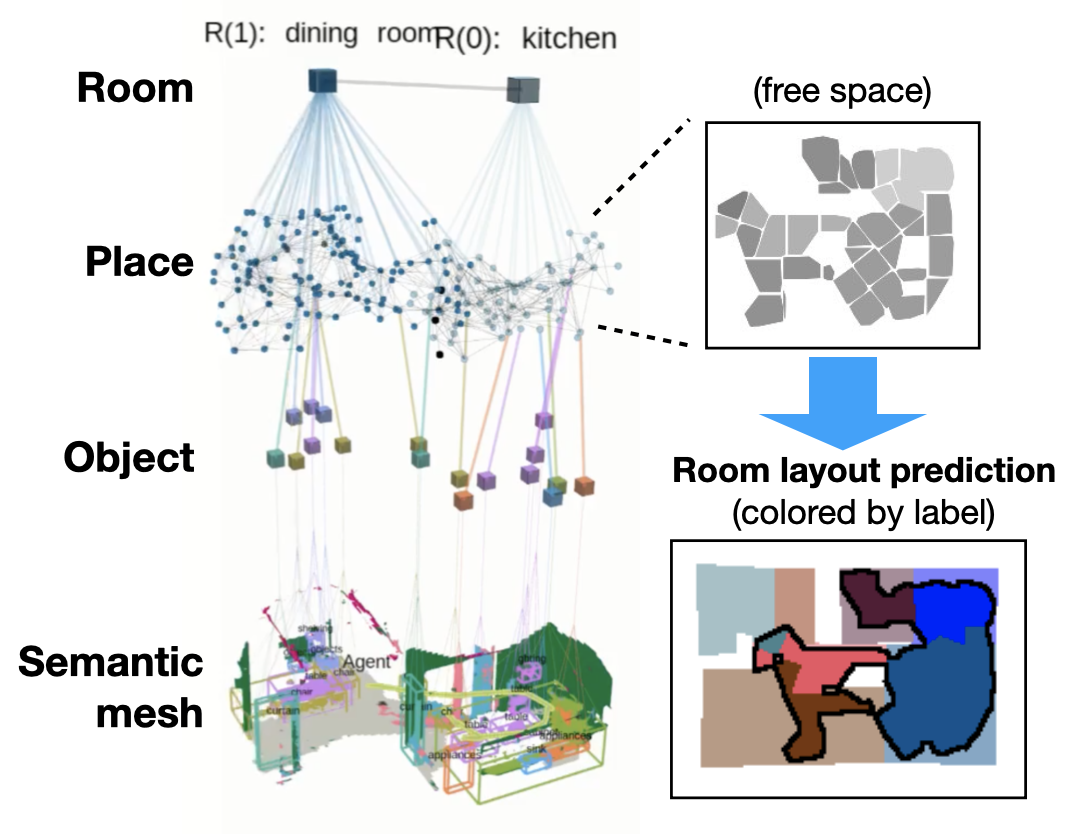}
        \caption{Partial room layout extraction via Hydra's place layer.\label{fig:hydra_processing_place}}
    \end{subfigure}\hfill
    \begin{subfigure}[t]{0.41\linewidth}
        \centering
        \includegraphics[width=\linewidth,trim=0 0 0 0mm,clip]{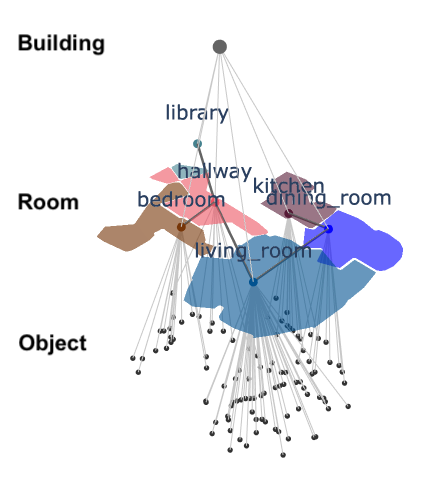}
        \caption{Partial building--room--object hierarchy extracted from Hydra outputs.\label{fig:hydra_processing_bro}}
    \end{subfigure}
    \caption{Hydra-based partial evidence extraction and construction of a partial building--room--object hierarchy.\label{fig:hydra_processing_overview}}
\end{figure}

\myParagraph{Extracting partial room geometry from the place layer}
Hydra constructs 3D scene graphs with multiple layers of abstraction from a semantic mesh built using RGB-D images and camera poses. For indoor scenes, Hydra builds a ``Place'' layer that represents free traversable spaces between the room and object layers.
For each room, we extract the boundary of the associated Place nodes as partial room inputs (\Cref{fig:hydra_processing_place}). We then pair these partial boundaries with the corresponding room labels (when available) and use them as constraints for our room-level scene graph generation module.
In addition, we connect room node pairs whose children Place nodes are directly connected.
\Cref{fig:hydra_processing_overview} provides an overview of the extracted partial room geometry and the resulting partial hierarchy.

\myParagraph{Fusing into a building-room-object hierarchy}
Hydra outputs object nodes as children of the place layer (\ie observed objects are attached to the closest free traversable space).
After we merge Place nodes into partial room geometry, we re-attach observed object nodes to the room layer using the intermediate Place nodes.
Finally, if not already present, we add a building node as the parent of all room nodes to form a consistent building--room--object hierarchy (\Cref{fig:hydra_processing_bro}).

%% file: tex/supplementary/additionalResults.tex

\subsection{Example Synthesized Layout Images\label{app:additionalImages}}

We render room-level layouts (\ie predicted floor plans) on a white background for quantitative image-level evaluations. The colors in the layout images correspond to different room categories, with similar semantic labels mapped to similar RGB colors. Note that we exclude room connectivity lines in these images as they affect the clarity of node-level details in the generated room scene graphs. 

We provide example results from different methods, as well as ground-truth layouts in \Cref{fig:3dfront_Holodeck} through \Cref{fig:mp3d_inputs}. To better display the consistency of the predicted results with respect to input partial scenes, these example figures contain two additional processing steps compared to the layout images used for quantitative evaluations in \Cref{sec:experiment}. 
First, we add the boundary of the input partial scene onto the layout images as a solid black outline. Second, we dilute the colors in the room areas with 0.6 alpha blending with a white background to better contrast the black boundary lines.

These additional results further support the conclusions in \Cref{sec:experiment} that (1) the \gls{llm}-based methods (AnyHome and Holodeck) struggle with conforming to input partial scenes and, without dataset-specific training, generate a different layout distribution than the data distribution of interest; (2) the occupancy-based method (SSC) carries a risk of dropping semantic constraints and can generate fine-grained and realistic layouts only given in-distribution inputs; (3) our method generates realistic layouts that are consistent with input partial scenes and generalize better to out-of-distribution inputs, with a small trade-off of slight overlap between rooms.

\input{figures/conditioned_3dfront/AnyHome/qualitative}
\clearpage
\input{figures/conditioned_3dfront/HoloDeck/qualitative}
\clearpage
\input{figures/conditioned_3dfront/SSC/qualitative}
\clearpage
\input{figures/conditioned_3dfront/Ours/qualitative}
\clearpage
\input{figures/conditioned_3dfront/GT/3dfront_inputs}
\clearpage

\input{figures/conditioned_mp3d/AnyHome/qualitative}
\clearpage
\input{figures/conditioned_mp3d/HoloDeck/qualitative}
\clearpage
\input{figures/conditioned_mp3d/SSC/qualitative}
\clearpage
\input{figures/conditioned_mp3d/Ours/qualitative}
\clearpage
\input{figures/conditioned_mp3d/GT/mp3d_inputs}
\clearpage

%% file: figures/conditioned_3dfront/AnyHome/qualitative.tex

\begin{figure*}[t]
    \centering

    \begin{subfigure}{\linewidth}
        \centering
        \fbox{\includegraphics[width=0.18\linewidth]{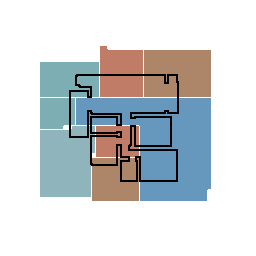}}
        \fbox{\includegraphics[width=0.18\linewidth]{figures/cross_256x256.png}}
        \fbox{\includegraphics[width=0.18\linewidth]{figures/cross_256x256.png}}
        \fbox{\includegraphics[width=0.18\linewidth]{figures/cross_256x256.png}}
        \fbox{\includegraphics[width=0.18\linewidth]{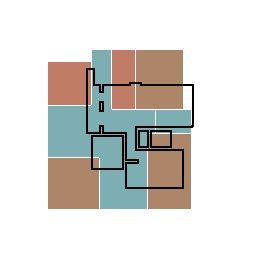}}
    \end{subfigure}

    \vspace{5pt}

    \begin{subfigure}{\linewidth}
        \centering
        \fbox{\includegraphics[width=0.18\linewidth]{figures/cross_256x256.png}}
        \fbox{\includegraphics[width=0.18\linewidth]{figures/cross_256x256.png}}
        \fbox{\includegraphics[width=0.18\linewidth]{figures/cross_256x256.png}}
        \fbox{\includegraphics[width=0.18\linewidth]{figures/cross_256x256.png}}
        \fbox{\includegraphics[width=0.18\linewidth]{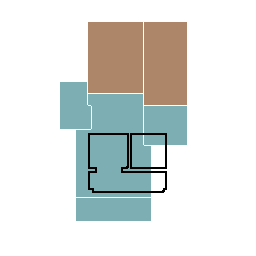}}
    \end{subfigure}

    \vspace{5pt}

    \begin{subfigure}{\linewidth}
        \centering
        \fbox{\includegraphics[width=0.18\linewidth]{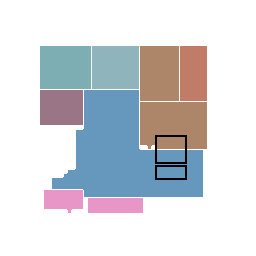}}
        \fbox{\includegraphics[width=0.18\linewidth]{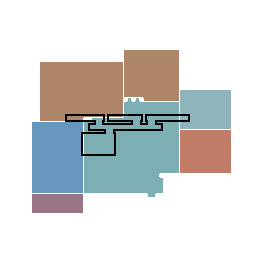}}
        \fbox{\includegraphics[width=0.18\linewidth]{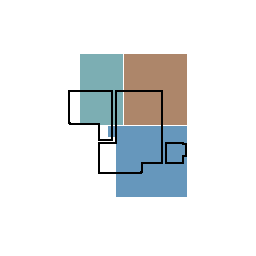}}
        \fbox{\includegraphics[width=0.18\linewidth]{figures/cross_256x256.png}}
        \fbox{\includegraphics[width=0.18\linewidth]{figures/conditioned_3dfront/AnyHome/blended_14.png}}
    \end{subfigure}

    \vspace{5pt}

    \begin{subfigure}{\linewidth}
        \centering
        \fbox{\includegraphics[width=0.18\linewidth]{figures/cross_256x256.png}}
        \fbox{\includegraphics[width=0.18\linewidth]{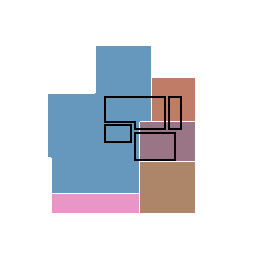}}
        \fbox{\includegraphics[width=0.18\linewidth]{figures/cross_256x256.png}}
        \fbox{\includegraphics[width=0.18\linewidth]{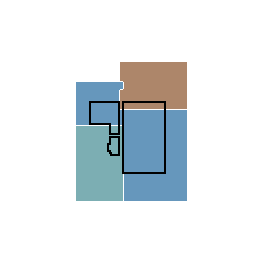}}
        \fbox{\includegraphics[width=0.18\linewidth]{figures/cross_256x256.png}}
    \end{subfigure}

    \vspace{5pt}
    
    \begin{subfigure}{\linewidth}
        \centering
        \fbox{\includegraphics[width=0.18\linewidth]{figures/cross_256x256.png}}
        \fbox{\includegraphics[width=0.18\linewidth]{figures/cross_256x256.png}}
        \fbox{\includegraphics[width=0.18\linewidth]{figures/cross_256x256.png}}
        \fbox{\includegraphics[width=0.18\linewidth]{figures/cross_256x256.png}}
        \fbox{\includegraphics[width=0.18\linewidth]{figures/cross_256x256.png}}
    \end{subfigure}

    \vspace{5pt}

    \begin{subfigure}{\linewidth}
        \centering
        \fbox{\includegraphics[width=0.18\linewidth]{figures/cross_256x256.png}}
        \fbox{\includegraphics[width=0.18\linewidth]{figures/cross_256x256.png}}
        \fbox{\includegraphics[width=0.18\linewidth]{figures/cross_256x256.png}}
        \fbox{\includegraphics[width=0.18\linewidth]{figures/cross_256x256.png}}
        \fbox{\includegraphics[width=0.18\linewidth]{figures/cross_256x256.png}}
    \end{subfigure}

    \caption{Example 3D-FRONT results using AnyHome. \label{fig:3dfront_anyhome}}
\end{figure*}

%% file: figures/conditioned_3dfront/HoloDeck/qualitative.tex

\begin{figure*}[t]
    \centering

    \begin{subfigure}{\linewidth}
        \centering
        \fbox{\includegraphics[width=0.18\linewidth]{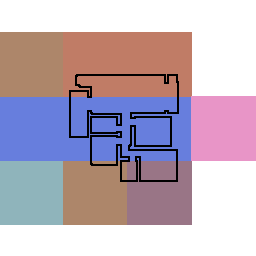}}
        \fbox{\includegraphics[width=0.18\linewidth]{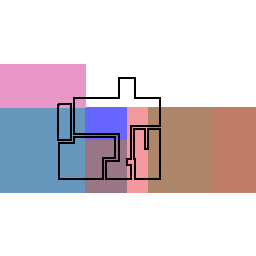}}
        \fbox{\includegraphics[width=0.18\linewidth]{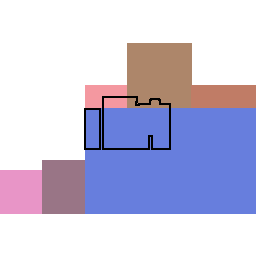}}
        \fbox{\includegraphics[width=0.18\linewidth]{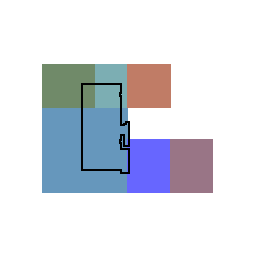}}
        \fbox{\includegraphics[width=0.18\linewidth]{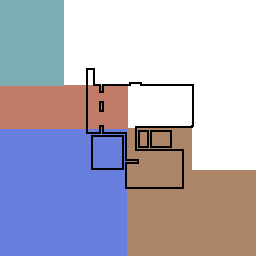}}
    \end{subfigure}

    \vspace{5pt}

    \begin{subfigure}{\linewidth}
        \centering
        \fbox{\includegraphics[width=0.18\linewidth]{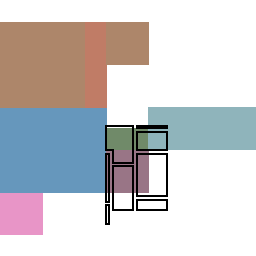}}
        \fbox{\includegraphics[width=0.18\linewidth]{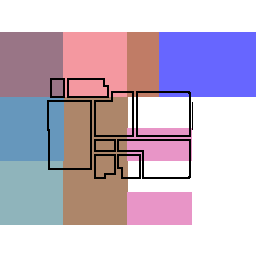}}
        \fbox{\includegraphics[width=0.18\linewidth]{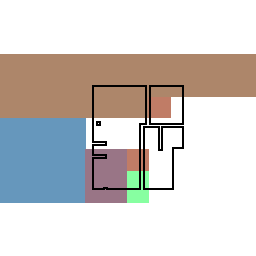}}
        \fbox{\includegraphics[width=0.18\linewidth]{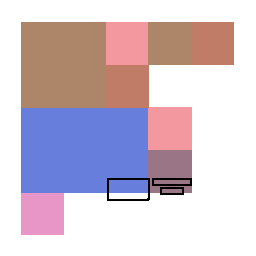}}
        \fbox{\includegraphics[width=0.18\linewidth]{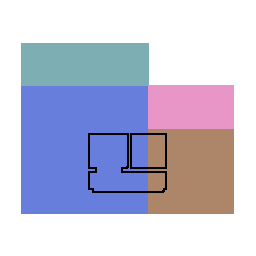}}
    \end{subfigure}

    \vspace{5pt}

    \begin{subfigure}{\linewidth}
        \centering
        \fbox{\includegraphics[width=0.18\linewidth]{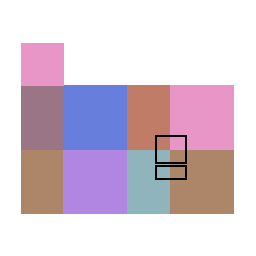}}
        \fbox{\includegraphics[width=0.18\linewidth]{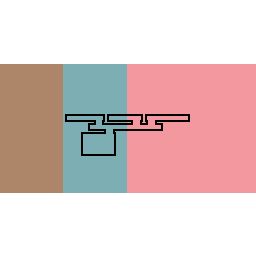}}
        \fbox{\includegraphics[width=0.18\linewidth]{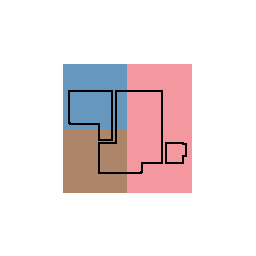}}
        \fbox{\includegraphics[width=0.18\linewidth]{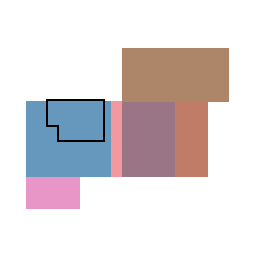}}
        \fbox{\includegraphics[width=0.18\linewidth]{figures/conditioned_3dfront/HoloDeck/blended_14.png}}
    \end{subfigure}

    \vspace{5pt}

    \begin{subfigure}{\linewidth}
        \centering
        \fbox{\includegraphics[width=0.18\linewidth]{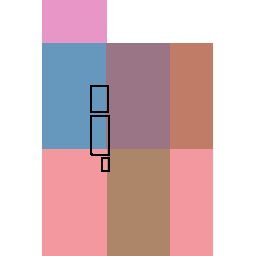}}
        \fbox{\includegraphics[width=0.18\linewidth]{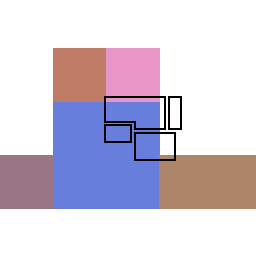}}
        \fbox{\includegraphics[width=0.18\linewidth]{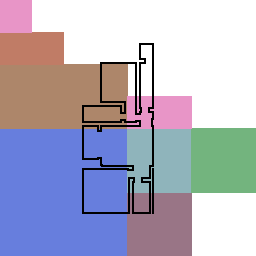}}
        \fbox{\includegraphics[width=0.18\linewidth]{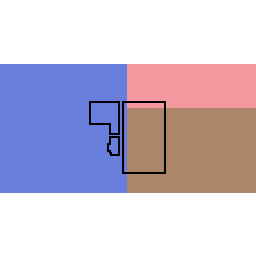}}
        \fbox{\includegraphics[width=0.18\linewidth]{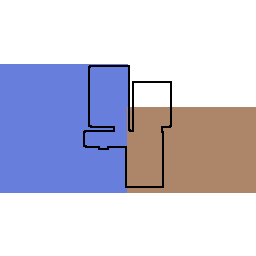}}
    \end{subfigure}

    \vspace{5pt}
    
    \begin{subfigure}{\linewidth}
        \centering
        \fbox{\includegraphics[width=0.18\linewidth]{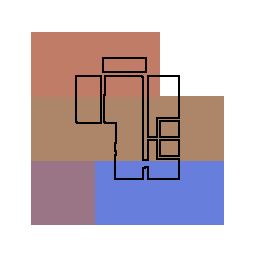}}
        \fbox{\includegraphics[width=0.18\linewidth]{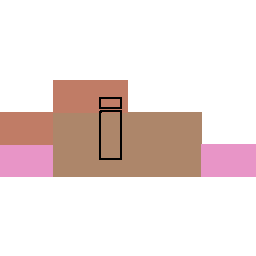}}
        \fbox{\includegraphics[width=0.18\linewidth]{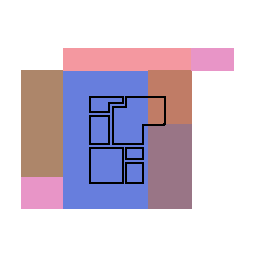}}
        \fbox{\includegraphics[width=0.18\linewidth]{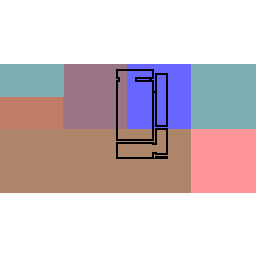}}
        \fbox{\includegraphics[width=0.18\linewidth]{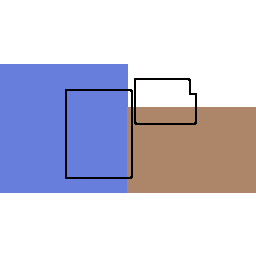}}
    \end{subfigure}

    \vspace{5pt}

    \begin{subfigure}{\linewidth}
        \centering
        \fbox{\includegraphics[width=0.18\linewidth]{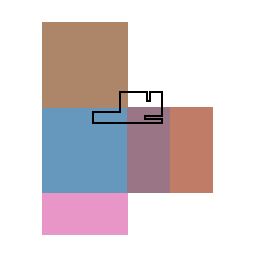}}
        \fbox{\includegraphics[width=0.18\linewidth]{figures/conditioned_3dfront/HoloDeck/blended_26.png}}
        \fbox{\includegraphics[width=0.18\linewidth]{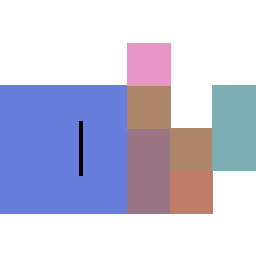}}
        \fbox{\includegraphics[width=0.18\linewidth]{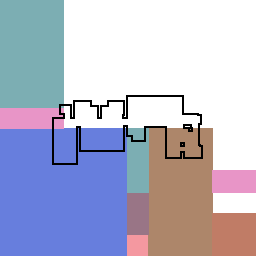}}
        \fbox{\includegraphics[width=0.18\linewidth]{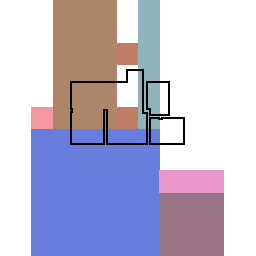}}
    \end{subfigure}

    \caption{Example 3D-FRONT results using Holodeck. \label{fig:3dfront_Holodeck}}
\end{figure*}

%% file: figures/conditioned_3dfront/SSC/qualitative.tex

\begin{figure*}[t]
    \centering

    \begin{subfigure}{\linewidth}
        \centering
        \fbox{\includegraphics[width=0.18\linewidth]{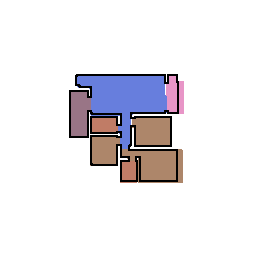}}
        \fbox{\includegraphics[width=0.18\linewidth]{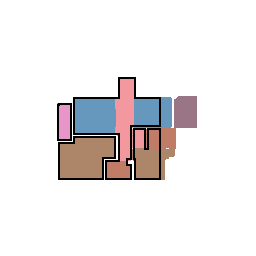}}
        \fbox{\includegraphics[width=0.18\linewidth]{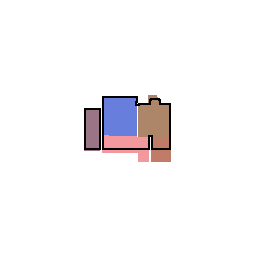}}
        \fbox{\includegraphics[width=0.18\linewidth]{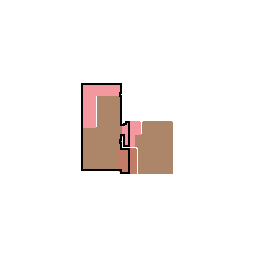}}
        \fbox{\includegraphics[width=0.18\linewidth]{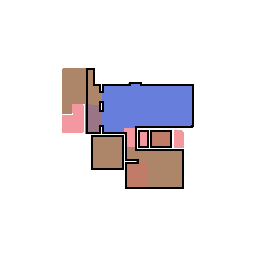}}
    \end{subfigure}

    \vspace{5pt}

    \begin{subfigure}{\linewidth}
        \centering
        \fbox{\includegraphics[width=0.18\linewidth]{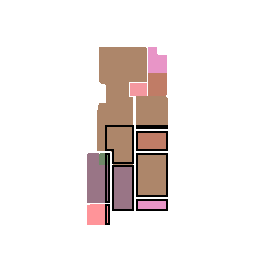}}
        \fbox{\includegraphics[width=0.18\linewidth]{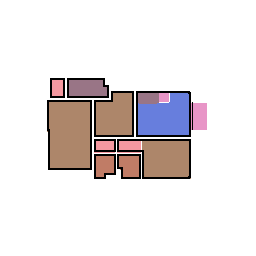}}
        \fbox{\includegraphics[width=0.18\linewidth]{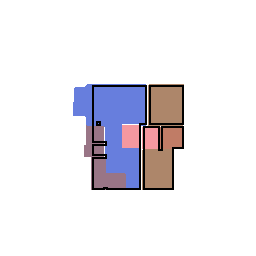}}
        \fbox{\includegraphics[width=0.18\linewidth]{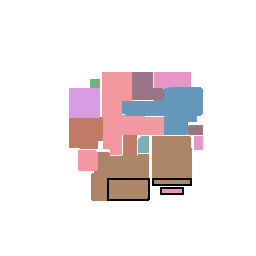}}
        \fbox{\includegraphics[width=0.18\linewidth]{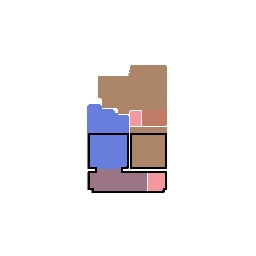}}
    \end{subfigure}

    \vspace{5pt}

    \begin{subfigure}{\linewidth}
        \centering
        \fbox{\includegraphics[width=0.18\linewidth]{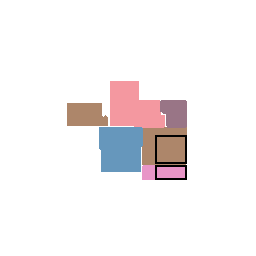}}
        \fbox{\includegraphics[width=0.18\linewidth]{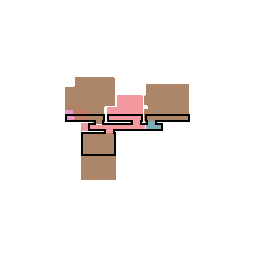}}
        \fbox{\includegraphics[width=0.18\linewidth]{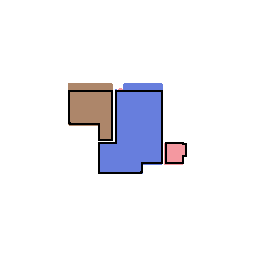}}
        \fbox{\includegraphics[width=0.18\linewidth]{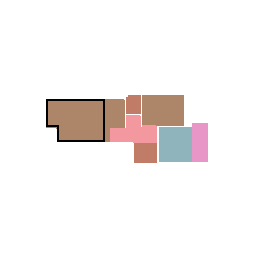}}
        \fbox{\includegraphics[width=0.18\linewidth]{figures/conditioned_3dfront/SSC/blended_14.png}}
    \end{subfigure}

    \vspace{5pt}

    \begin{subfigure}{\linewidth}
        \centering
        \fbox{\includegraphics[width=0.18\linewidth]{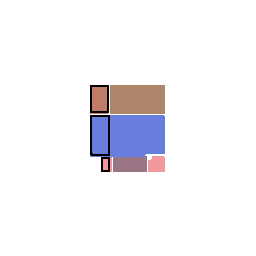}}
        \fbox{\includegraphics[width=0.18\linewidth]{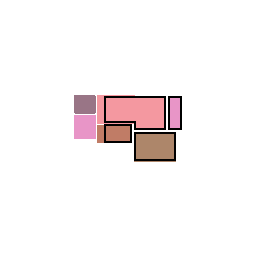}}
        \fbox{\includegraphics[width=0.18\linewidth]{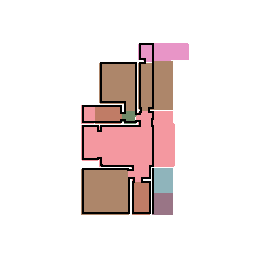}}
        \fbox{\includegraphics[width=0.18\linewidth]{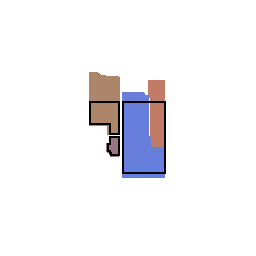}}
        \fbox{\includegraphics[width=0.18\linewidth]{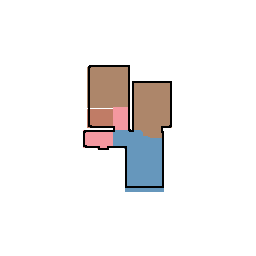}}
    \end{subfigure}

    \vspace{5pt}
    
    \begin{subfigure}{\linewidth}
        \centering
        \fbox{\includegraphics[width=0.18\linewidth]{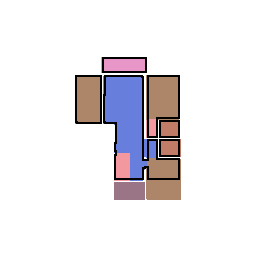}}
        \fbox{\includegraphics[width=0.18\linewidth]{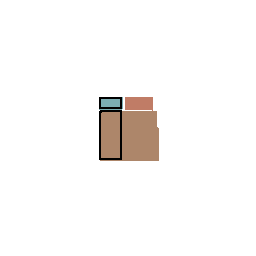}}
        \fbox{\includegraphics[width=0.18\linewidth]{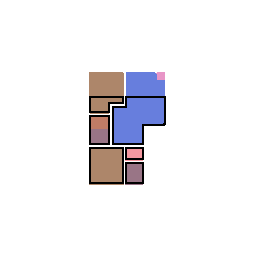}}
        \fbox{\includegraphics[width=0.18\linewidth]{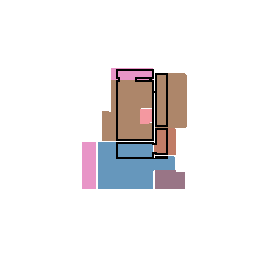}}
        \fbox{\includegraphics[width=0.18\linewidth]{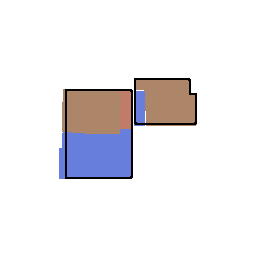}}
    \end{subfigure}

    \vspace{5pt}

    \begin{subfigure}{\linewidth}
        \centering
        \fbox{\includegraphics[width=0.18\linewidth]{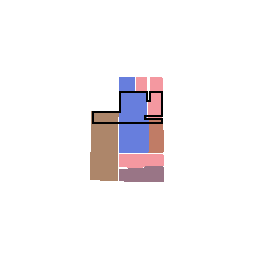}}
        \fbox{\includegraphics[width=0.18\linewidth]{figures/conditioned_3dfront/SSC/blended_26.png}}
        \fbox{\includegraphics[width=0.18\linewidth]{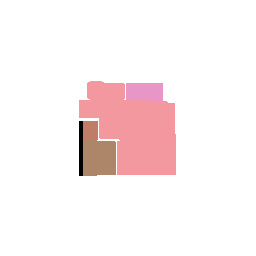}}
        \fbox{\includegraphics[width=0.18\linewidth]{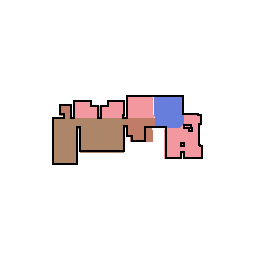}}
        \fbox{\includegraphics[width=0.18\linewidth]{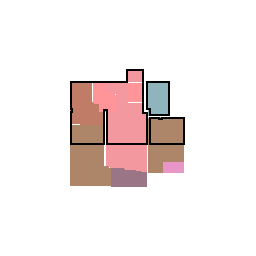}}
    \end{subfigure}
    
    \caption{Example 3D-FRONT results using SSC. \label{fig:3dfront_ssc}}
\end{figure*}

%% file: figures/conditioned_3dfront/Ours/qualitative.tex

\begin{figure*}[t]
    \centering

    \begin{subfigure}{\linewidth}
        \centering
        \fbox{\includegraphics[width=0.18\linewidth]{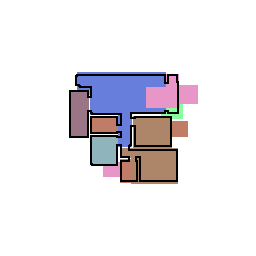}}
        \fbox{\includegraphics[width=0.18\linewidth]{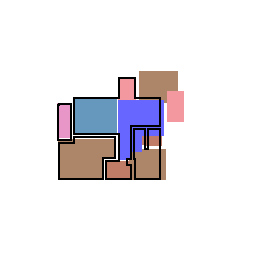}}
        \fbox{\includegraphics[width=0.18\linewidth]{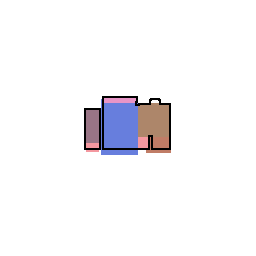}}
        \fbox{\includegraphics[width=0.18\linewidth]{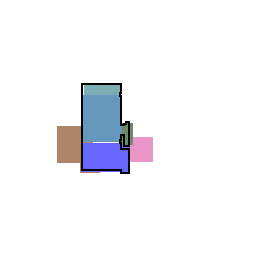}}
        \fbox{\includegraphics[width=0.18\linewidth]{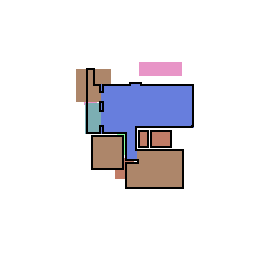}}
    \end{subfigure}

    \vspace{5pt}

    \begin{subfigure}{\linewidth}
        \centering
        \fbox{\includegraphics[width=0.18\linewidth]{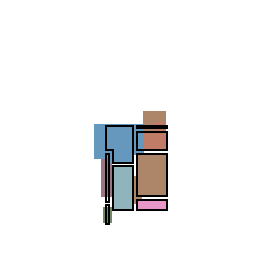}}
        \fbox{\includegraphics[width=0.18\linewidth]{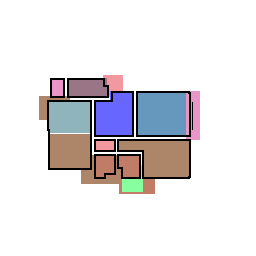}}
        \fbox{\includegraphics[width=0.18\linewidth]{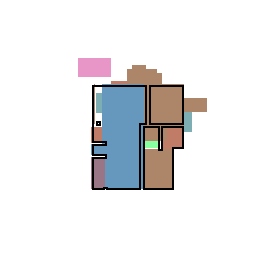}}
        \fbox{\includegraphics[width=0.18\linewidth]{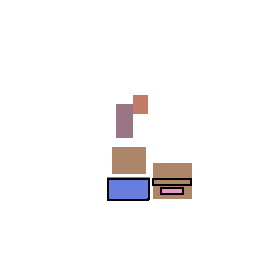}}
        \fbox{\includegraphics[width=0.18\linewidth]{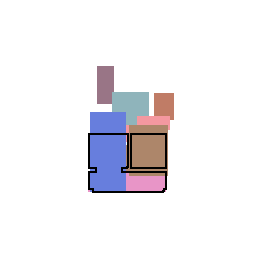}}
    \end{subfigure}

    \vspace{5pt}

    \begin{subfigure}{\linewidth}
        \centering
        \fbox{\includegraphics[width=0.18\linewidth]{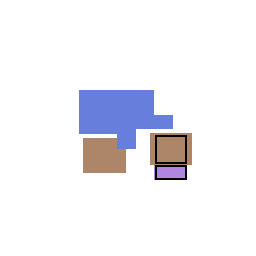}}
        \fbox{\includegraphics[width=0.18\linewidth]{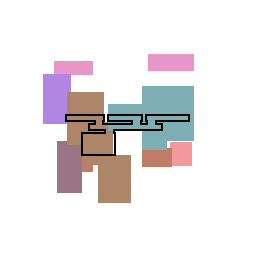}}
        \fbox{\includegraphics[width=0.18\linewidth]{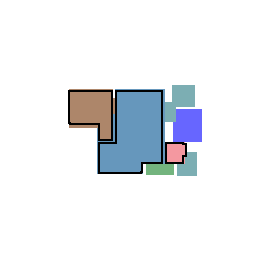}}
        \fbox{\includegraphics[width=0.18\linewidth]{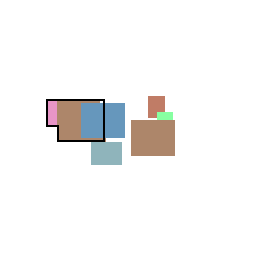}}
        \fbox{\includegraphics[width=0.18\linewidth]{figures/conditioned_3dfront/Ours/blended_14.png}}
    \end{subfigure}

    \vspace{5pt}

    \begin{subfigure}{\linewidth}
        \centering
        \fbox{\includegraphics[width=0.18\linewidth]{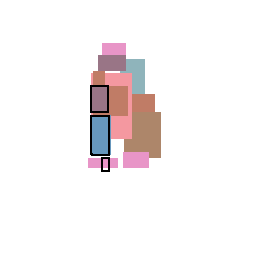}}
        \fbox{\includegraphics[width=0.18\linewidth]{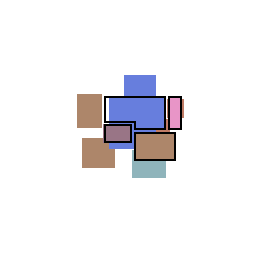}}
        \fbox{\includegraphics[width=0.18\linewidth]{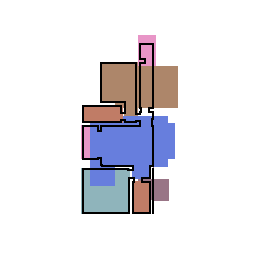}}
        \fbox{\includegraphics[width=0.18\linewidth]{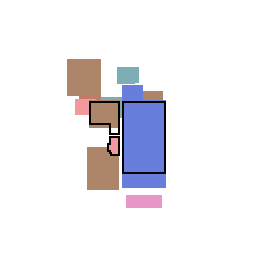}}
        \fbox{\includegraphics[width=0.18\linewidth]{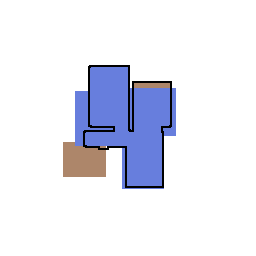}}
    \end{subfigure}

    \vspace{5pt}
    
    \begin{subfigure}{\linewidth}
        \centering
        \fbox{\includegraphics[width=0.18\linewidth]{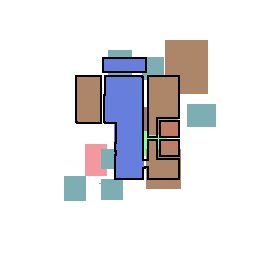}}
        \fbox{\includegraphics[width=0.18\linewidth]{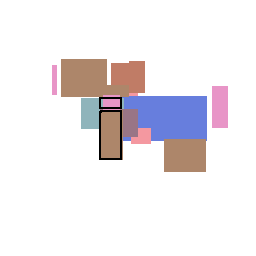}}
        \fbox{\includegraphics[width=0.18\linewidth]{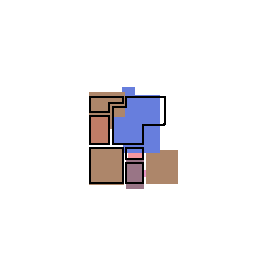}}
        \fbox{\includegraphics[width=0.18\linewidth]{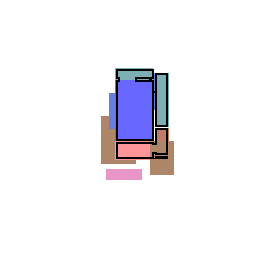}}
        \fbox{\includegraphics[width=0.18\linewidth]{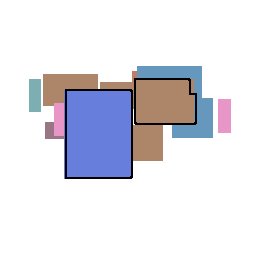}}
    \end{subfigure}

    \vspace{5pt}

    \begin{subfigure}{\linewidth}
        \centering
        \fbox{\includegraphics[width=0.18\linewidth]{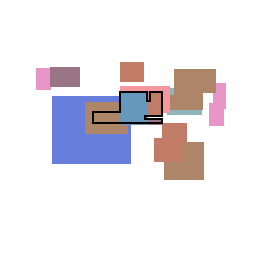}}
        \fbox{\includegraphics[width=0.18\linewidth]{figures/conditioned_3dfront/Ours/blended_26.png}}
        \fbox{\includegraphics[width=0.18\linewidth]{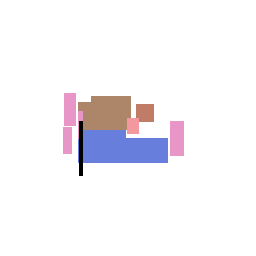}}
        \fbox{\includegraphics[width=0.18\linewidth]{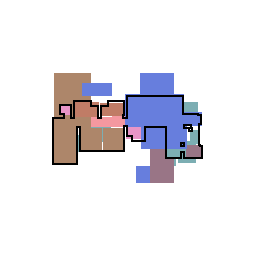}}
        \fbox{\includegraphics[width=0.18\linewidth]{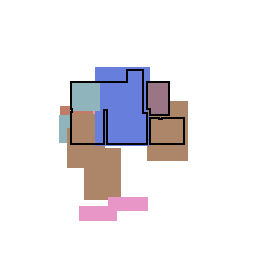}}
    \end{subfigure}

    \caption{Example 3D-FRONT results using our method. \label{fig:3dfront_ours}}
\end{figure*}

%% file: figures/conditioned_3dfront/GT/3dfront_inputs.tex

\begin{figure*}[t]
    \centering

    \begin{subfigure}{\linewidth}
        \centering
        \fbox{\includegraphics[width=0.18\linewidth]{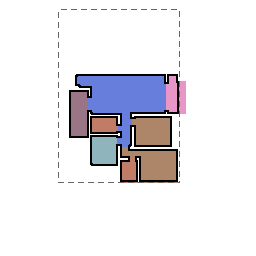}}
        \fbox{\includegraphics[width=0.18\linewidth]{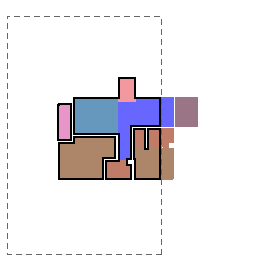}}
        \fbox{\includegraphics[width=0.18\linewidth]{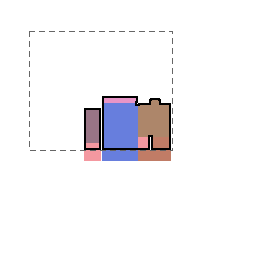}}
        \fbox{\includegraphics[width=0.18\linewidth]{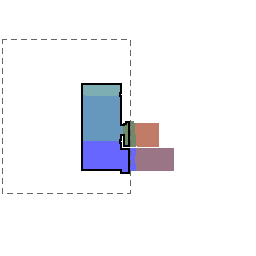}}
        \fbox{\includegraphics[width=0.18\linewidth]{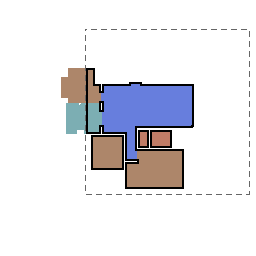}}
    \end{subfigure}

    \vspace{5pt}

    \begin{subfigure}{\linewidth}
        \centering
        \fbox{\includegraphics[width=0.18\linewidth]{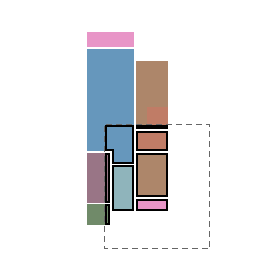}}
        \fbox{\includegraphics[width=0.18\linewidth]{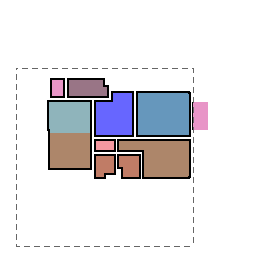}}
        \fbox{\includegraphics[width=0.18\linewidth]{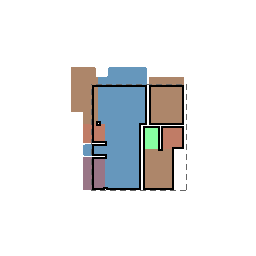}}
        \fbox{\includegraphics[width=0.18\linewidth]{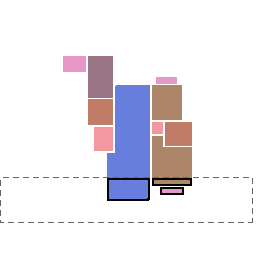}}
        \fbox{\includegraphics[width=0.18\linewidth]{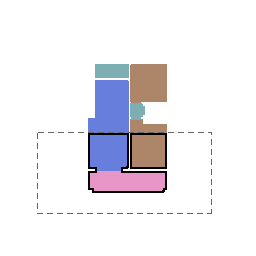}}
    \end{subfigure}

    \vspace{5pt}

    \begin{subfigure}{\linewidth}
        \centering
        \fbox{\includegraphics[width=0.18\linewidth]{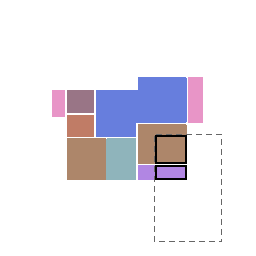}}
        \fbox{\includegraphics[width=0.18\linewidth]{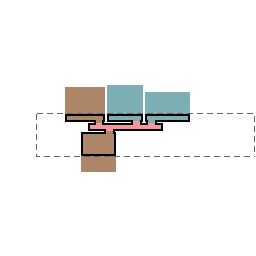}}
        \fbox{\includegraphics[width=0.18\linewidth]{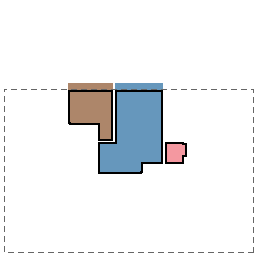}}
        \fbox{\includegraphics[width=0.18\linewidth]{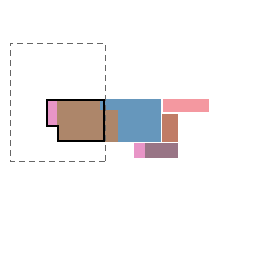}}
        \fbox{\includegraphics[width=0.18\linewidth]{figures/conditioned_3dfront/GT/blended_14.png}}
    \end{subfigure}

    \vspace{5pt}

    \begin{subfigure}{\linewidth}
        \centering
        \fbox{\includegraphics[width=0.18\linewidth]{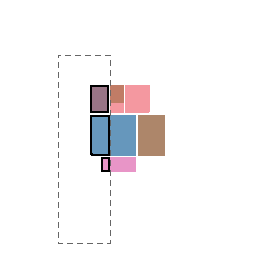}}
        \fbox{\includegraphics[width=0.18\linewidth]{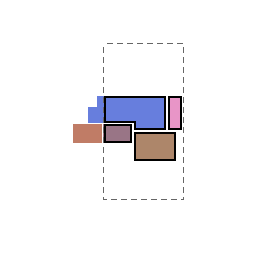}}
        \fbox{\includegraphics[width=0.18\linewidth]{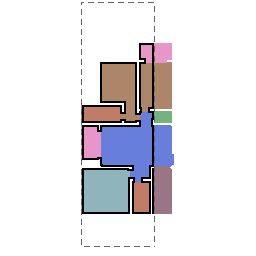}}
        \fbox{\includegraphics[width=0.18\linewidth]{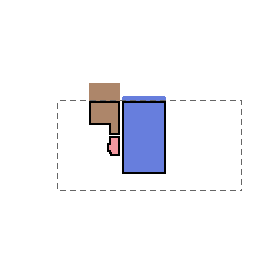}}
        \fbox{\includegraphics[width=0.18\linewidth]{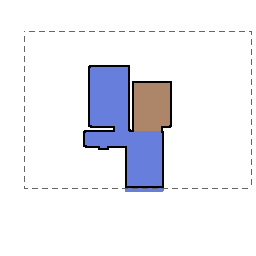}}
    \end{subfigure}

    \vspace{5pt}
    
    \begin{subfigure}{\linewidth}
        \centering
        \fbox{\includegraphics[width=0.18\linewidth]{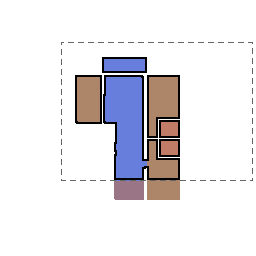}}
        \fbox{\includegraphics[width=0.18\linewidth]{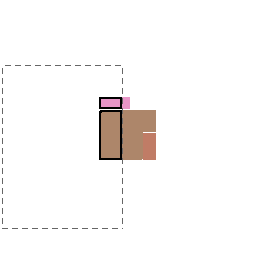}}
        \fbox{\includegraphics[width=0.18\linewidth]{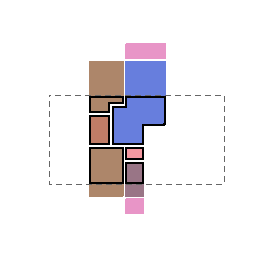}}
        \fbox{\includegraphics[width=0.18\linewidth]{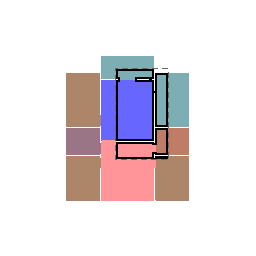}}
        \fbox{\includegraphics[width=0.18\linewidth]{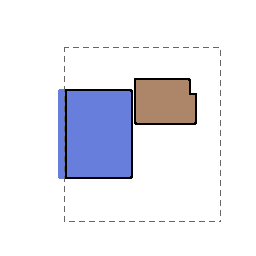}}
    \end{subfigure}

    \vspace{5pt}

    \begin{subfigure}{\linewidth}
        \centering
        \fbox{\includegraphics[width=0.18\linewidth]{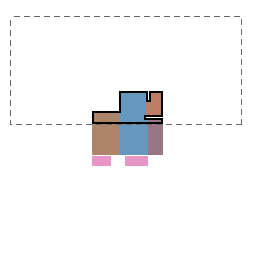}}
        \fbox{\includegraphics[width=0.18\linewidth]{figures/conditioned_3dfront/GT/blended_26.png}}
        \fbox{\includegraphics[width=0.18\linewidth]{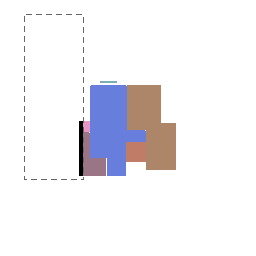}}
        \fbox{\includegraphics[width=0.18\linewidth]{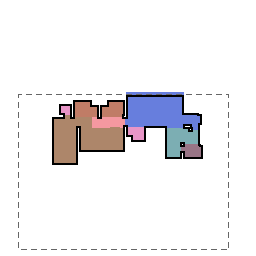}}
        \fbox{\includegraphics[width=0.18\linewidth]{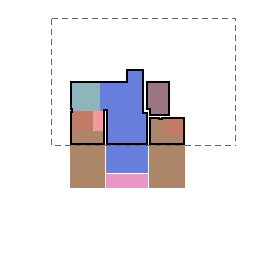}}
    \end{subfigure}
    
    \caption{3D-FRONT input partial scenes. \label{fig:3dfront_inputs}}
\end{figure*}

%% file: figures/conditioned_mp3d/AnyHome/qualitative.tex

\begin{figure*}[t]
    \centering

    \begin{subfigure}{\linewidth}
        \centering
        \fbox{\includegraphics[width=0.18\linewidth]{figures/cross_256x256.png}}%
        \fbox{\includegraphics[width=0.18\linewidth]{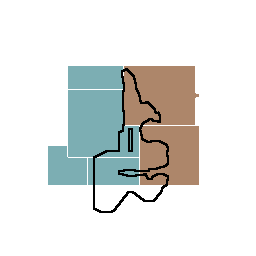}}%
        \fbox{\includegraphics[width=0.18\linewidth]{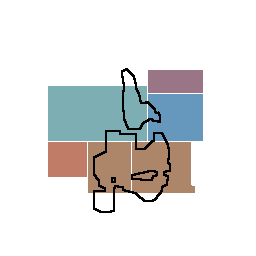}}%
        \fbox{\includegraphics[width=0.18\linewidth]{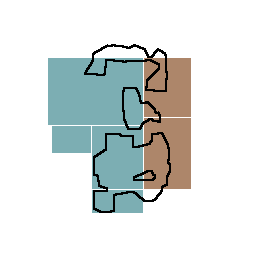}}%
        \fbox{\includegraphics[width=0.18\linewidth]{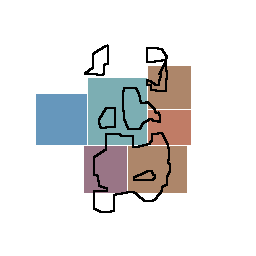}}%
        \vspace{-5pt}
        \caption{17DRP5sb8fy}
        \vspace{3pt}
    \end{subfigure}

    \begin{subfigure}{\linewidth}
        \centering
        \fbox{\includegraphics[width=0.18\linewidth]{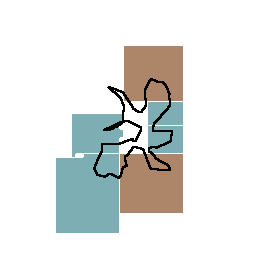}}%
        \fbox{\includegraphics[width=0.18\linewidth]{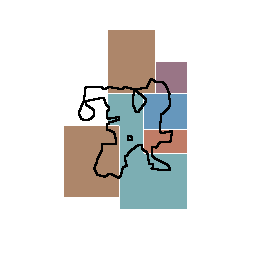}}%
        \fbox{\includegraphics[width=0.18\linewidth]{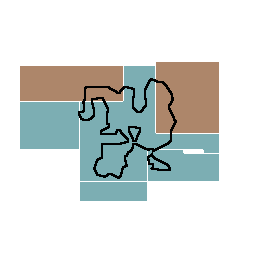}}%
        \fbox{\includegraphics[width=0.18\linewidth]{figures/cross_256x256.png}}%
        \fbox{\includegraphics[width=0.18\linewidth]{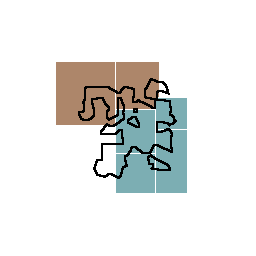}}%
        \vspace{-5pt}
        \caption{2t7WUuJeko7}
        \vspace{3pt}
    \end{subfigure}

    \begin{subfigure}{\linewidth}
        \centering
        \fbox{\includegraphics[width=0.18\linewidth]{figures/cross_256x256.png}}%
        \fbox{\includegraphics[width=0.18\linewidth]{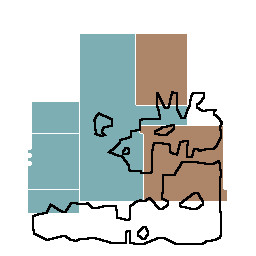}}%
        \fbox{\includegraphics[width=0.18\linewidth]{figures/cross_256x256.png}}%
        \fbox{\includegraphics[width=0.18\linewidth]{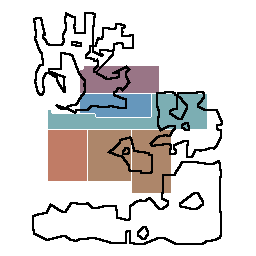}}%
        \fbox{\includegraphics[width=0.18\linewidth]{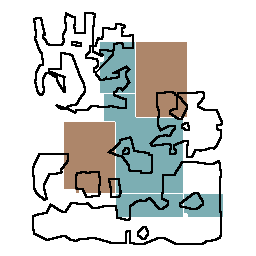}}%
        \vspace{-5pt}
        \caption{JeFG25nYj2p}
        \vspace{3pt}
    \end{subfigure}

    \begin{subfigure}{\linewidth}
        \centering
        \fbox{\includegraphics[width=0.18\linewidth]{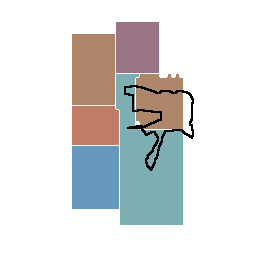}}%
        \fbox{\includegraphics[width=0.18\linewidth]{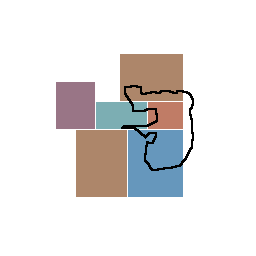}}%
        \fbox{\includegraphics[width=0.18\linewidth]{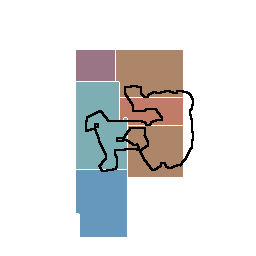}}%
        \fbox{\includegraphics[width=0.18\linewidth]{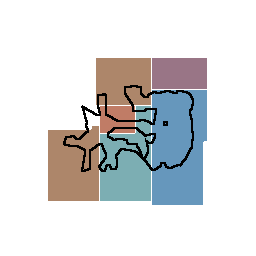}}%
        \fbox{\includegraphics[width=0.18\linewidth]{figures/conditioned_mp3d/AnyHome/blended_225.png}}%
        \vspace{-5pt}
        \caption{RPmz2sHmrrY}
        \vspace{3pt}
    \end{subfigure}
    
    \begin{subfigure}{\linewidth}
        \centering
        \fbox{\includegraphics[width=0.18\linewidth]{figures/cross_256x256.png}}%
        \fbox{\includegraphics[width=0.18\linewidth]{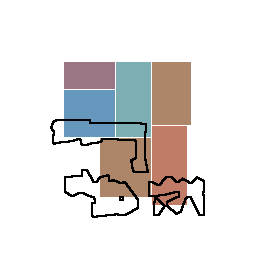}}%
        \fbox{\includegraphics[width=0.18\linewidth]{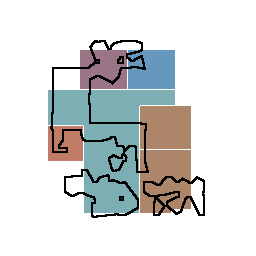}}%
        \fbox{\includegraphics[width=0.18\linewidth]{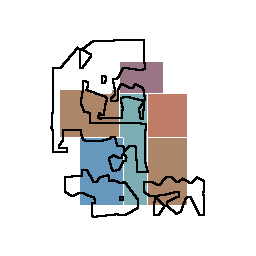}}%
        \fbox{\includegraphics[width=0.18\linewidth]{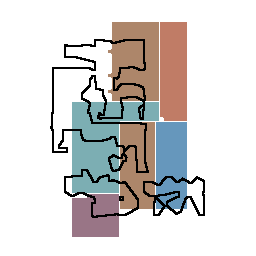}}%
        \vspace{-5pt}
        \caption{jh4fc5c5qoQ}
        \vspace{3pt}
    \end{subfigure}

    \begin{subfigure}{\linewidth}
        \centering
        \fbox{\includegraphics[width=0.18\linewidth]{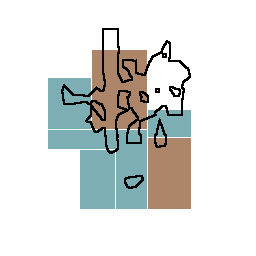}}%
        \fbox{\includegraphics[width=0.18\linewidth]{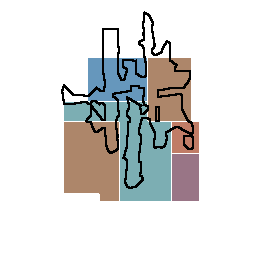}}%
        \fbox{\includegraphics[width=0.18\linewidth]{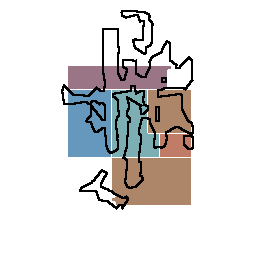}}%
        \fbox{\includegraphics[width=0.18\linewidth]{figures/cross_256x256.png}}%
        \fbox{\includegraphics[width=0.18\linewidth]{figures/conditioned_mp3d/AnyHome/blended_235.png}}%
        \vspace{-5pt}
        \caption{zsNo4HB9uLZ}
        \vspace{3pt}
    \end{subfigure}
    
    \vspace{-10pt}
    \caption{Example MP3D results using AnyHome. \label{fig:mp3d_anyhome}}
\end{figure*}

%% file: figures/conditioned_mp3d/HoloDeck/qualitative.tex

\begin{figure*}[t]
    \centering

    \begin{subfigure}{\linewidth}
        \centering
        \fbox{\includegraphics[width=0.18\linewidth]{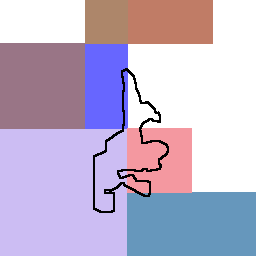}}%
        \fbox{\includegraphics[width=0.18\linewidth]{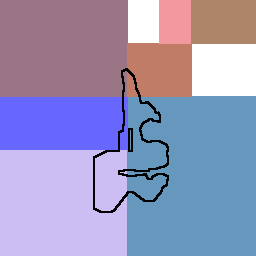}}%
        \fbox{\includegraphics[width=0.18\linewidth]{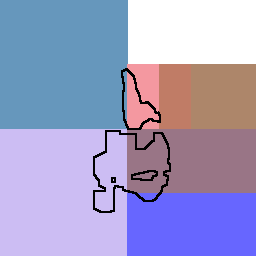}}%
        \fbox{\includegraphics[width=0.18\linewidth]{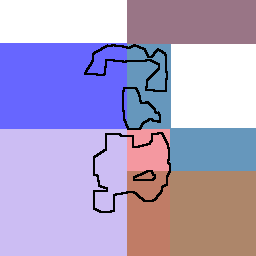}}%
        \fbox{\includegraphics[width=0.18\linewidth]{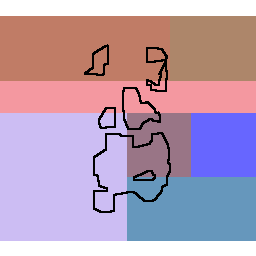}}%
        \vspace{-5pt}
        \caption{17DRP5sb8fy}
        \vspace{3pt}
    \end{subfigure}

    \begin{subfigure}{\linewidth}
        \centering
        \fbox{\includegraphics[width=0.18\linewidth]{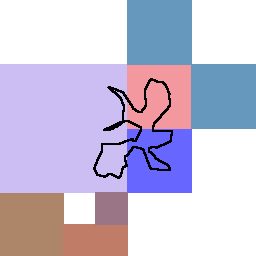}}%
        \fbox{\includegraphics[width=0.18\linewidth]{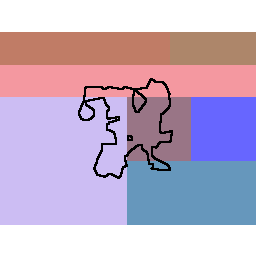}}%
        \fbox{\includegraphics[width=0.18\linewidth]{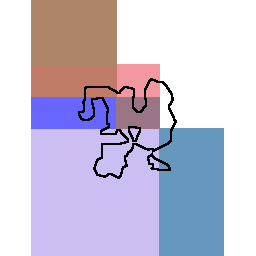}}%
        \fbox{\includegraphics[width=0.18\linewidth]{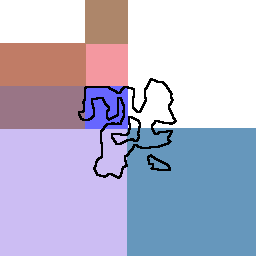}}%
        \fbox{\includegraphics[width=0.18\linewidth]{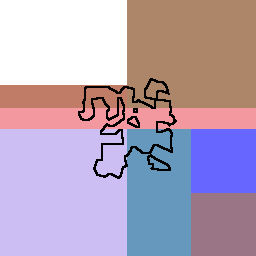}}%
        \vspace{-5pt}
        \caption{2t7WUuJeko7}
        \vspace{3pt}
    \end{subfigure}

    \begin{subfigure}{\linewidth}
        \centering
        \fbox{\includegraphics[width=0.18\linewidth]{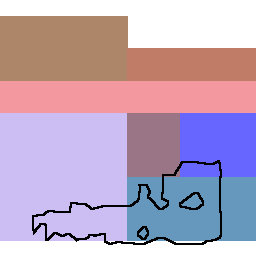}}%
        \fbox{\includegraphics[width=0.18\linewidth]{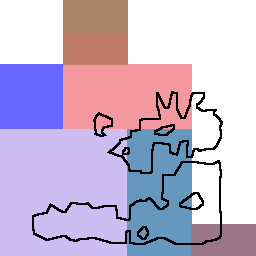}}%
        \fbox{\includegraphics[width=0.18\linewidth]{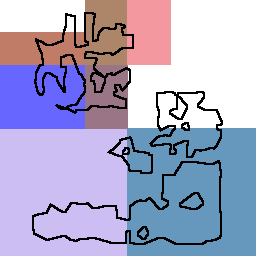}}%
        \fbox{\includegraphics[width=0.18\linewidth]{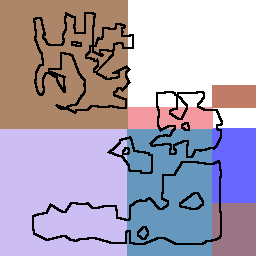}}%
        \fbox{\includegraphics[width=0.18\linewidth]{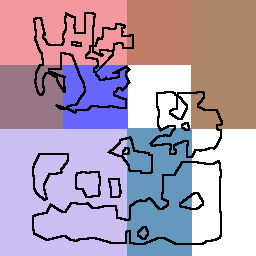}}%
        \vspace{-5pt}
        \caption{JeFG25nYj2p}
        \vspace{3pt}
    \end{subfigure}

    \begin{subfigure}{\linewidth}
        \centering
        \fbox{\includegraphics[width=0.18\linewidth]{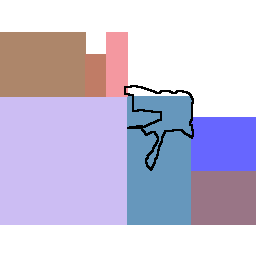}}%
        \fbox{\includegraphics[width=0.18\linewidth]{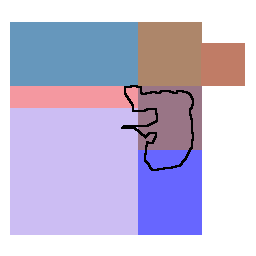}}%
        \fbox{\includegraphics[width=0.18\linewidth]{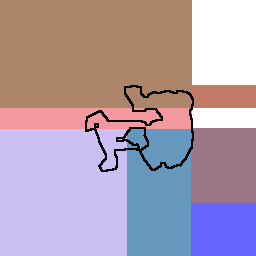}}%
        \fbox{\includegraphics[width=0.18\linewidth]{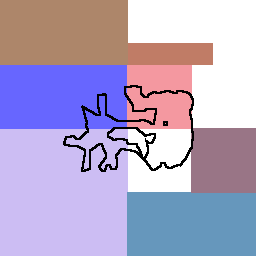}}%
        \fbox{\includegraphics[width=0.18\linewidth]{figures/conditioned_mp3d/HoloDeck/blended_15.png}}%
        \vspace{-5pt}
        \caption{RPmz2sHmrrY}
        \vspace{3pt}
    \end{subfigure}
    
    \begin{subfigure}{\linewidth}
        \centering
        \fbox{\includegraphics[width=0.18\linewidth]{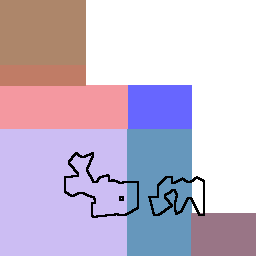}}%
        \fbox{\includegraphics[width=0.18\linewidth]{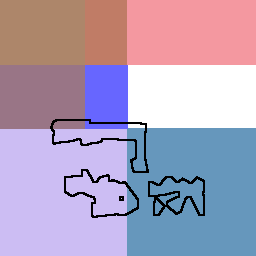}}%
        \fbox{\includegraphics[width=0.18\linewidth]{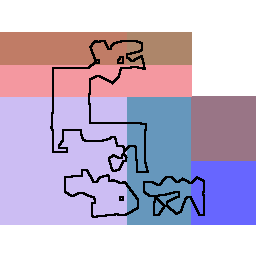}}%
        \fbox{\includegraphics[width=0.18\linewidth]{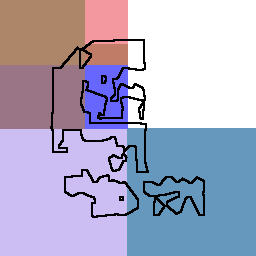}}%
        \fbox{\includegraphics[width=0.18\linewidth]{figures/conditioned_mp3d/HoloDeck/blended_20.png}}%
        \vspace{-5pt}
        \caption{jh4fc5c5qoQ}
        \vspace{3pt}
    \end{subfigure}

    \begin{subfigure}{\linewidth}
        \centering
        \fbox{\includegraphics[width=0.18\linewidth]{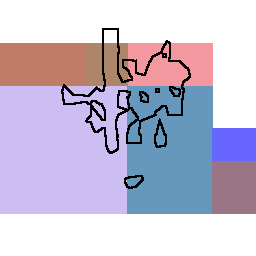}}%
        \fbox{\includegraphics[width=0.18\linewidth]{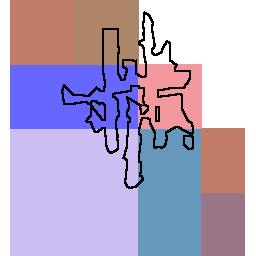}}%
        \fbox{\includegraphics[width=0.18\linewidth]{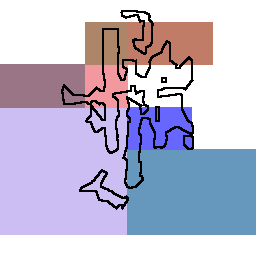}}%
        \fbox{\includegraphics[width=0.18\linewidth]{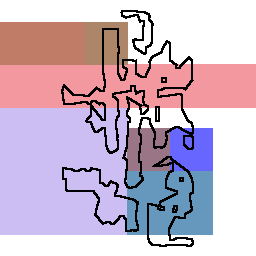}}%
        \fbox{\includegraphics[width=0.18\linewidth]{figures/conditioned_mp3d/HoloDeck/blended_25.png}}%
        \vspace{-5pt}
        \caption{zsNo4HB9uLZ}
        \vspace{3pt}
    \end{subfigure}
    
    \vspace{-10pt}
    \caption{Example MP3D results using Holodeck. \label{fig:mp3d_holodeck}}
\end{figure*}

%% file: figures/conditioned_mp3d/SSC/qualitative.tex

\begin{figure*}[t]
    \centering

    \begin{subfigure}{\linewidth}
        \centering
        \fbox{\includegraphics[width=0.18\linewidth]{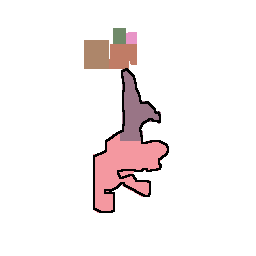}}%
        \fbox{\includegraphics[width=0.18\linewidth]{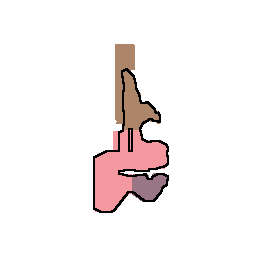}}%
        \fbox{\includegraphics[width=0.18\linewidth]{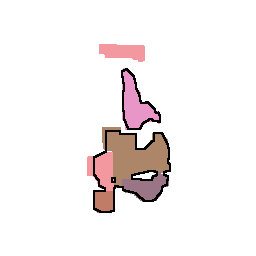}}%
        \fbox{\includegraphics[width=0.18\linewidth]{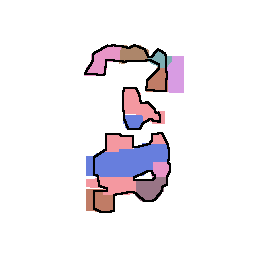}}%
        \fbox{\includegraphics[width=0.18\linewidth]{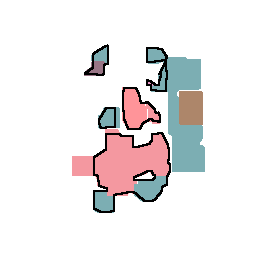}}%
        \vspace{-5pt}
        \caption{17DRP5sb8fy}
        \vspace{3pt}
    \end{subfigure}

    \begin{subfigure}{\linewidth}
        \centering
        \fbox{\includegraphics[width=0.18\linewidth]{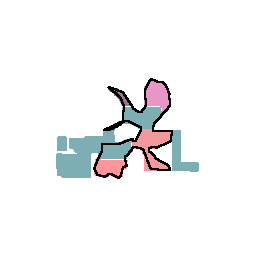}}%
        \fbox{\includegraphics[width=0.18\linewidth]{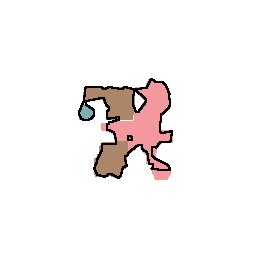}}%
        \fbox{\includegraphics[width=0.18\linewidth]{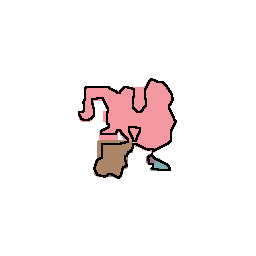}}%
        \fbox{\includegraphics[width=0.18\linewidth]{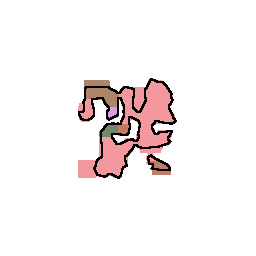}}%
        \fbox{\includegraphics[width=0.18\linewidth]{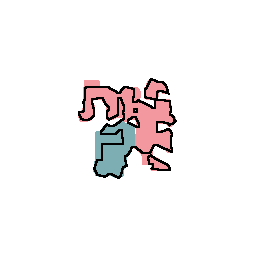}}%
        \vspace{-5pt}
        \caption{2t7WUuJeko7}
        \vspace{3pt}
    \end{subfigure}

    \begin{subfigure}{\linewidth}
        \centering
        \fbox{\includegraphics[width=0.18\linewidth]{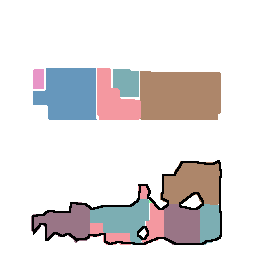}}%
        \fbox{\includegraphics[width=0.18\linewidth]{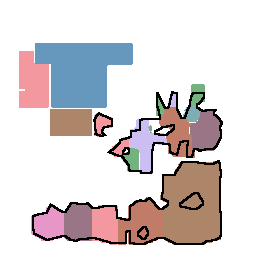}}%
        \fbox{\includegraphics[width=0.18\linewidth]{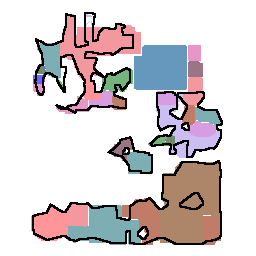}}%
        \fbox{\includegraphics[width=0.18\linewidth]{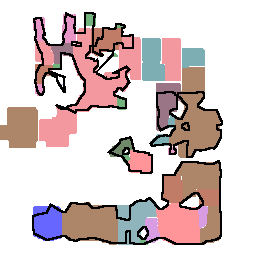}}%
        \fbox{\includegraphics[width=0.18\linewidth]{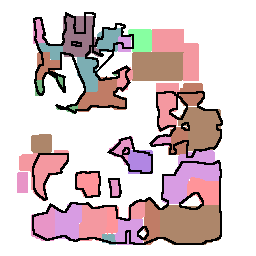}}%
        \vspace{-5pt}
        \caption{JeFG25nYj2p}
        \vspace{3pt}
    \end{subfigure}

    \begin{subfigure}{\linewidth}
        \centering
        \fbox{\includegraphics[width=0.18\linewidth]{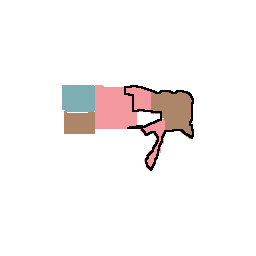}}%
        \fbox{\includegraphics[width=0.18\linewidth]{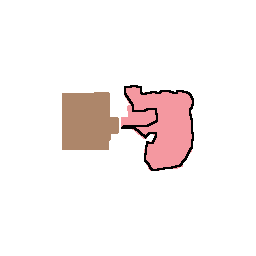}}%
        \fbox{\includegraphics[width=0.18\linewidth]{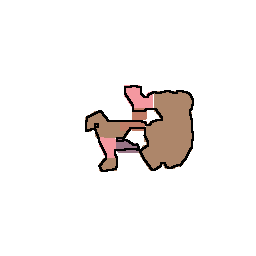}}%
        \fbox{\includegraphics[width=0.18\linewidth]{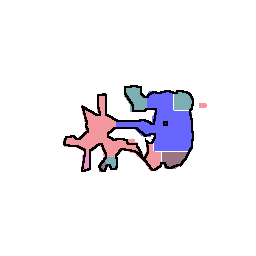}}%
        \fbox{\includegraphics[width=0.18\linewidth]{figures/conditioned_mp3d/SSC/blended_15.png}}%
        \vspace{-5pt}
        \caption{RPmz2sHmrrY}
        \vspace{3pt}
    \end{subfigure}
    
    \begin{subfigure}{\linewidth}
        \centering
        \fbox{\includegraphics[width=0.18\linewidth]{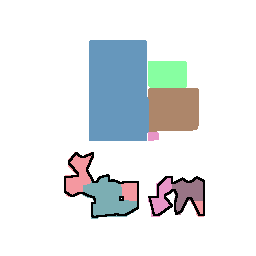}}%
        \fbox{\includegraphics[width=0.18\linewidth]{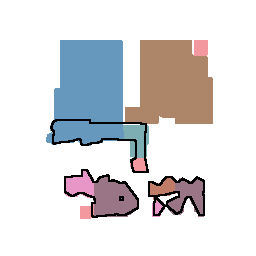}}%
        \fbox{\includegraphics[width=0.18\linewidth]{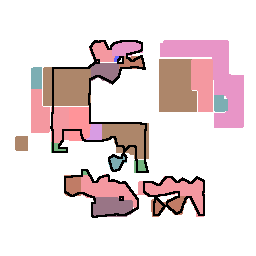}}%
        \fbox{\includegraphics[width=0.18\linewidth]{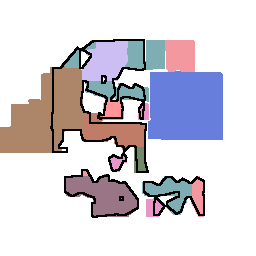}}%
        \fbox{\includegraphics[width=0.18\linewidth]{figures/conditioned_mp3d/SSC/blended_20.png}}%
        \vspace{-5pt}
        \caption{jh4fc5c5qoQ}
        \vspace{3pt}
    \end{subfigure}

    \begin{subfigure}{\linewidth}
        \centering
        \fbox{\includegraphics[width=0.18\linewidth]{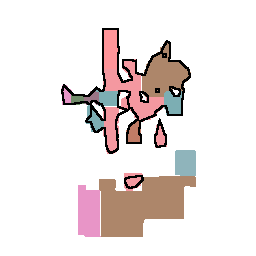}}%
        \fbox{\includegraphics[width=0.18\linewidth]{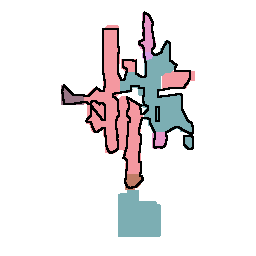}}%
        \fbox{\includegraphics[width=0.18\linewidth]{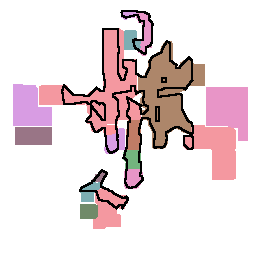}}%
        \fbox{\includegraphics[width=0.18\linewidth]{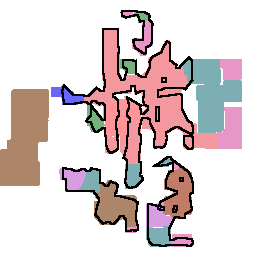}}%
        \fbox{\includegraphics[width=0.18\linewidth]{figures/conditioned_mp3d/SSC/blended_25.png}}%
        \vspace{-5pt}
        \caption{zsNo4HB9uLZ}
        \vspace{3pt}
    \end{subfigure}
    
    \vspace{-10pt}
    \caption{Example MP3D results using SSC. \label{fig:mp3d_ssc}}
\end{figure*}

%% file: figures/conditioned_mp3d/Ours/qualitative.tex

\begin{figure*}[t]
    \centering

    \begin{subfigure}{\linewidth}
        \centering
        \fbox{\includegraphics[width=0.18\linewidth]{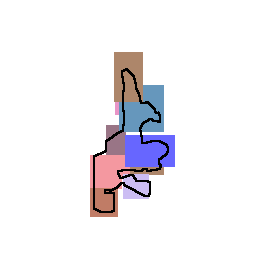}}%
        \fbox{\includegraphics[width=0.18\linewidth]{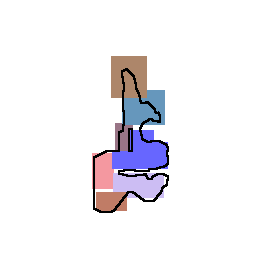}}%
        \fbox{\includegraphics[width=0.18\linewidth]{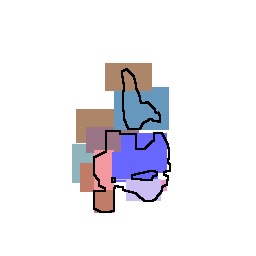}}%
        \fbox{\includegraphics[width=0.18\linewidth]{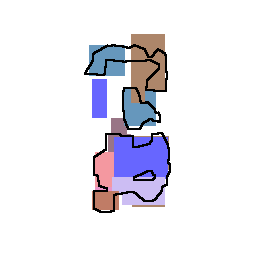}}%
        \fbox{\includegraphics[width=0.18\linewidth]{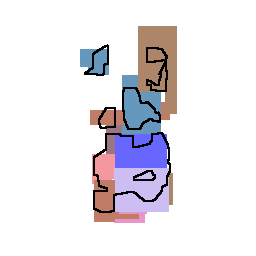}}%
        \vspace{-5pt}
        \caption{17DRP5sb8fy}
        \vspace{3pt}
    \end{subfigure}

    \begin{subfigure}{\linewidth}
        \centering
        \fbox{\includegraphics[width=0.18\linewidth]{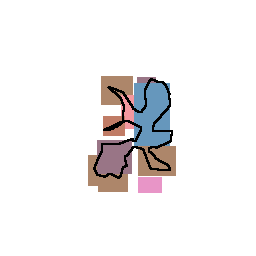}}%
        \fbox{\includegraphics[width=0.18\linewidth]{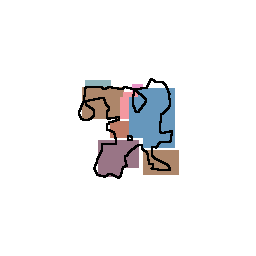}}%
        \fbox{\includegraphics[width=0.18\linewidth]{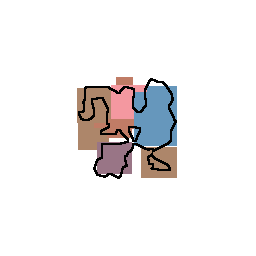}}%
        \fbox{\includegraphics[width=0.18\linewidth]{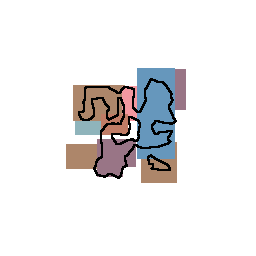}}%
        \fbox{\includegraphics[width=0.18\linewidth]{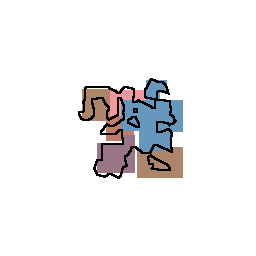}}%
        \vspace{-5pt}
        \caption{2t7WUuJeko7}
        \vspace{3pt}
    \end{subfigure}

    \begin{subfigure}{\linewidth}
        \centering
        \fbox{\includegraphics[width=0.18\linewidth]{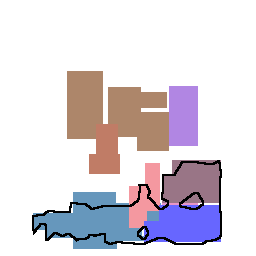}}%
        \fbox{\includegraphics[width=0.18\linewidth]{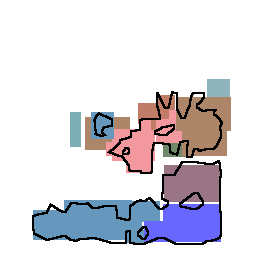}}%
        \fbox{\includegraphics[width=0.18\linewidth]{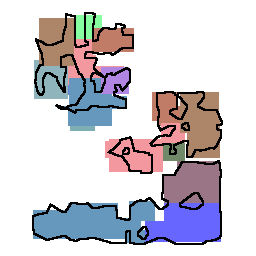}}%
        \fbox{\includegraphics[width=0.18\linewidth]{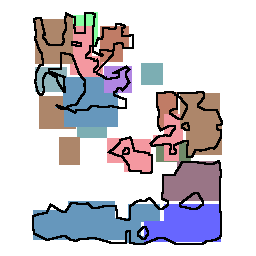}}%
        \fbox{\includegraphics[width=0.18\linewidth]{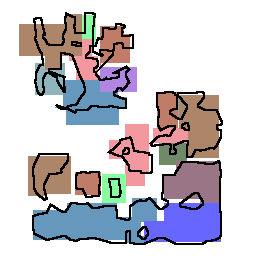}}%
        \vspace{-5pt}
        \caption{JeFG25nYj2p}
        \vspace{3pt}
    \end{subfigure}

    \begin{subfigure}{\linewidth}
        \centering
        \fbox{\includegraphics[width=0.18\linewidth]{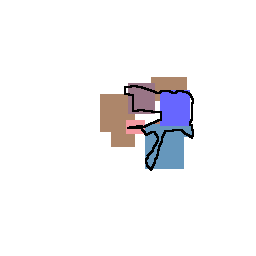}}%
        \fbox{\includegraphics[width=0.18\linewidth]{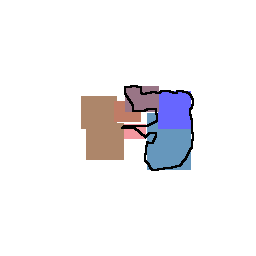}}%
        \fbox{\includegraphics[width=0.18\linewidth]{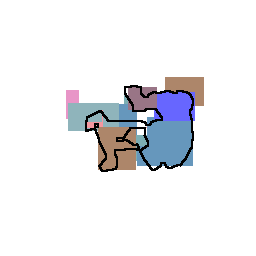}}%
        \fbox{\includegraphics[width=0.18\linewidth]{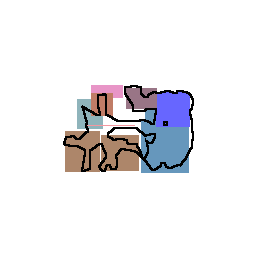}}%
        \fbox{\includegraphics[width=0.18\linewidth]{figures/conditioned_mp3d/Ours/blended_15.png}}%
        \vspace{-5pt}
        \caption{RPmz2sHmrrY}
        \vspace{3pt}
    \end{subfigure}
    
    \begin{subfigure}{\linewidth}
        \centering
        \fbox{\includegraphics[width=0.18\linewidth]{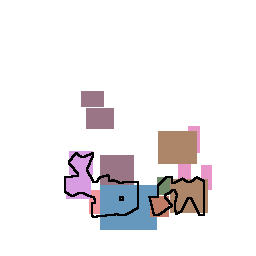}}%
        \fbox{\includegraphics[width=0.18\linewidth]{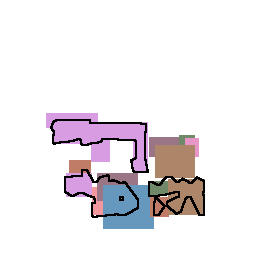}}%
        \fbox{\includegraphics[width=0.18\linewidth]{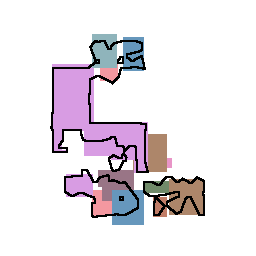}}%
        \fbox{\includegraphics[width=0.18\linewidth]{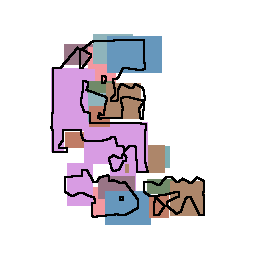}}%
        \fbox{\includegraphics[width=0.18\linewidth]{figures/conditioned_mp3d/Ours/blended_20.png}}%
        \vspace{-5pt}
        \caption{jh4fc5c5qoQ}
        \vspace{3pt}
    \end{subfigure}

    \begin{subfigure}{\linewidth}
        \centering
        \fbox{\includegraphics[width=0.18\linewidth]{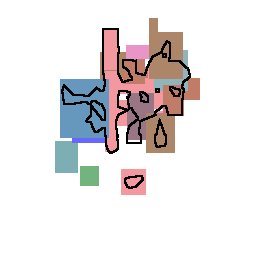}}%
        \fbox{\includegraphics[width=0.18\linewidth]{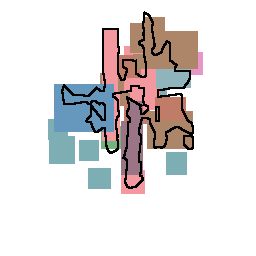}}%
        \fbox{\includegraphics[width=0.18\linewidth]{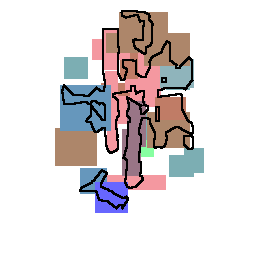}}%
        \fbox{\includegraphics[width=0.18\linewidth]{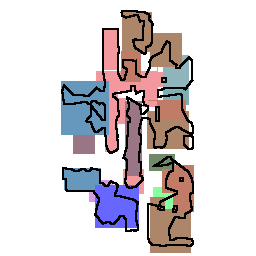}}%
        \fbox{\includegraphics[width=0.18\linewidth]{figures/conditioned_mp3d/Ours/blended_25.png}}%
        \vspace{-5pt}
        \caption{zsNo4HB9uLZ}
        \vspace{3pt}
    \end{subfigure}
    
    \vspace{-10pt}
    \caption{Example MP3D results using our method. \label{fig:mp3d_ours}}
\end{figure*}

%% file: figures/conditioned_mp3d/GT/mp3d_inputs.tex

\begin{figure*}[t]
    \centering

    \begin{subfigure}{\linewidth}
        \centering
        \fbox{\includegraphics[width=0.18\linewidth]{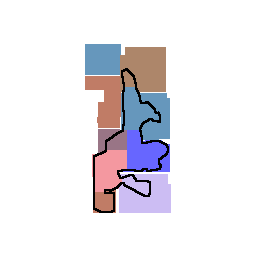}}%
        \fbox{\includegraphics[width=0.18\linewidth]{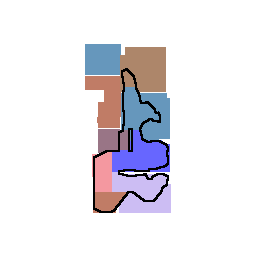}}%
        \fbox{\includegraphics[width=0.18\linewidth]{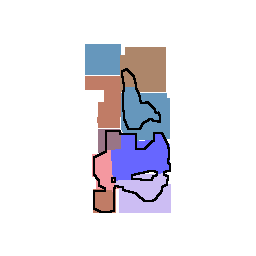}}%
        \fbox{\includegraphics[width=0.18\linewidth]{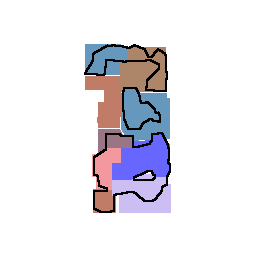}}%
        \fbox{\includegraphics[width=0.18\linewidth]{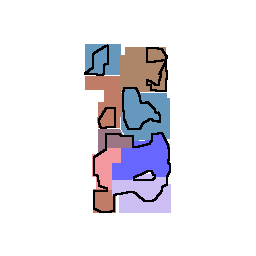}}%
        \vspace{-5pt}
        \caption{17DRP5sb8fy}
        \vspace{3pt}
    \end{subfigure}

    \begin{subfigure}{\linewidth}
        \centering
        \fbox{\includegraphics[width=0.18\linewidth]{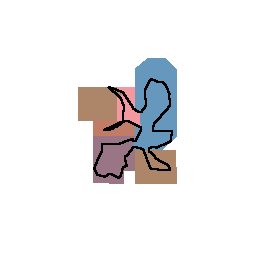}}%
        \fbox{\includegraphics[width=0.18\linewidth]{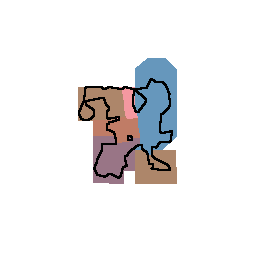}}%
        \fbox{\includegraphics[width=0.18\linewidth]{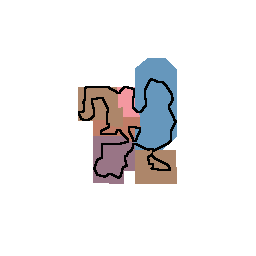}}%
        \fbox{\includegraphics[width=0.18\linewidth]{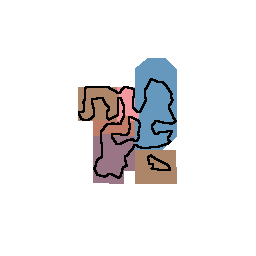}}%
        \fbox{\includegraphics[width=0.18\linewidth]{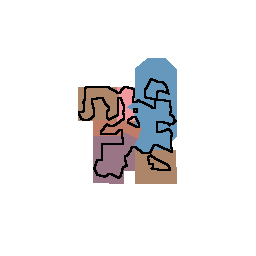}}%
        \vspace{-5pt}
        \caption{2t7WUuJeko7}
        \vspace{3pt}
    \end{subfigure}

    \begin{subfigure}{\linewidth}
        \centering
        \fbox{\includegraphics[width=0.18\linewidth]{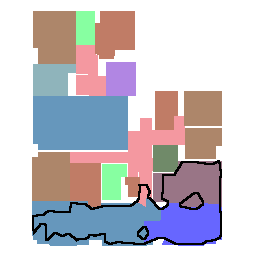}}%
        \fbox{\includegraphics[width=0.18\linewidth]{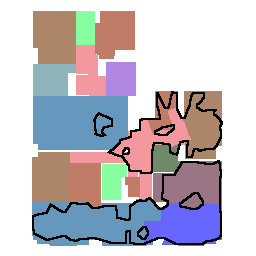}}%
        \fbox{\includegraphics[width=0.18\linewidth]{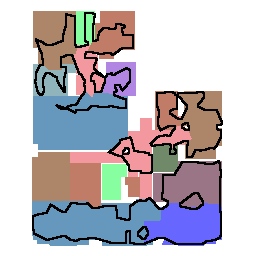}}%
        \fbox{\includegraphics[width=0.18\linewidth]{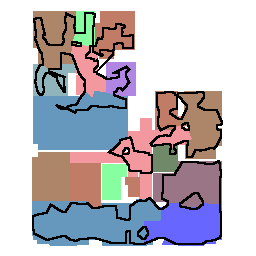}}%
        \fbox{\includegraphics[width=0.18\linewidth]{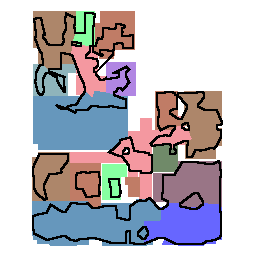}}%
        \vspace{-5pt}
        \caption{JeFG25nYj2p}
        \vspace{3pt}
    \end{subfigure}

    \begin{subfigure}{\linewidth}
        \centering
        \fbox{\includegraphics[width=0.18\linewidth]{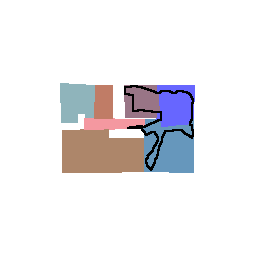}}%
        \fbox{\includegraphics[width=0.18\linewidth]{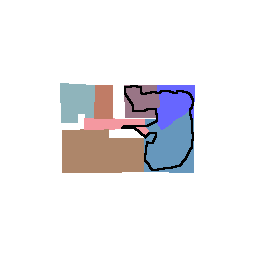}}%
        \fbox{\includegraphics[width=0.18\linewidth]{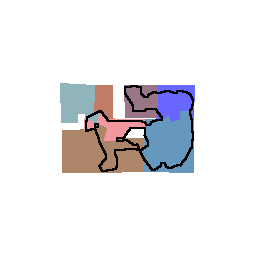}}%
        \fbox{\includegraphics[width=0.18\linewidth]{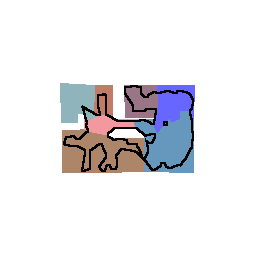}}%
        \fbox{\includegraphics[width=0.18\linewidth]{figures/conditioned_mp3d/GT/blended_15.png}}%
        \vspace{-5pt}
        \caption{RPmz2sHmrrY}
        \vspace{3pt}
    \end{subfigure}
    
    \begin{subfigure}{\linewidth}
        \centering
        \fbox{\includegraphics[width=0.18\linewidth]{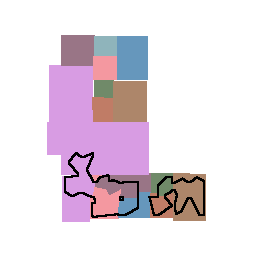}}%
        \fbox{\includegraphics[width=0.18\linewidth]{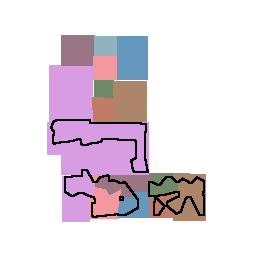}}%
        \fbox{\includegraphics[width=0.18\linewidth]{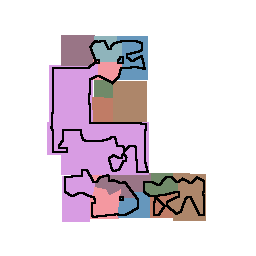}}%
        \fbox{\includegraphics[width=0.18\linewidth]{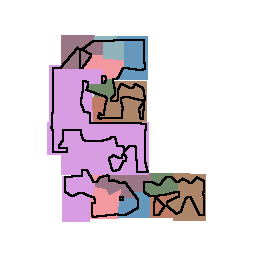}}%
        \fbox{\includegraphics[width=0.18\linewidth]{figures/conditioned_mp3d/GT/blended_20.png}}%
        \vspace{-5pt}
        \caption{jh4fc5c5qoQ}
        \vspace{3pt}
    \end{subfigure}

    \begin{subfigure}{\linewidth}
        \centering
        \fbox{\includegraphics[width=0.18\linewidth]{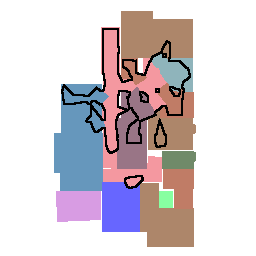}}%
        \fbox{\includegraphics[width=0.18\linewidth]{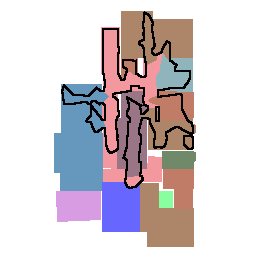}}%
        \fbox{\includegraphics[width=0.18\linewidth]{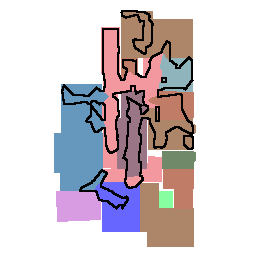}}%
        \fbox{\includegraphics[width=0.18\linewidth]{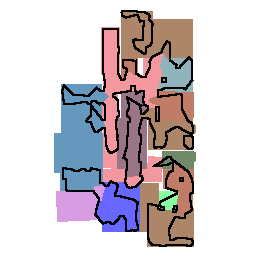}}%
        \fbox{\includegraphics[width=0.18\linewidth]{figures/conditioned_mp3d/GT/blended_25.png}}%
        \vspace{-5pt}
        \caption{zsNo4HB9uLZ}
        \vspace{3pt}
    \end{subfigure}
    
    \vspace{-10pt}
    \caption{MP3D input partial scenes. \label{fig:mp3d_inputs}}
\end{figure*}